\documentclass{article}

\PassOptionsToPackage{numbers}{natbib}

\usepackage[final]{neurips_data_2023}

\usepackage[utf8]{inputenc} % allow utf-8 input
\usepackage[T1]{fontenc}    % use 8-bit T1 fonts
\usepackage{hyperref}       % hyperlinks
\usepackage{url}            % simple URL typesetting
\usepackage{booktabs}       % professional-quality tables
\usepackage{amsfonts}       % blackboard math symbols
\usepackage{nicefrac}       % compact symbols for 1/2, etc.
\usepackage{microtype}      % microtypography
\usepackage{xcolor}         % colors
\usepackage{graphicx}
\usepackage{lipsum}
\usepackage{todonotes} % use this for notes and todos
\usepackage[]{markdown}
\usepackage{booktabs}
\usepackage{listings}
\usepackage{amsmath}
\usepackage{csvsimple}
\usepackage{float}
\usepackage{wrapfig}
\usepackage{placeins}
\newcommand*\red{\color{red}} % human outlines
\newcommand*\blue{\color{blue}} % long-form text, maybe generated, to be reviewed
\newcommand*\git{\url{https://github.com/LAION-AI/Open-Assistant}}

\newcommand*\oacontributors{\url{https://open-assistant.io/contributors}}
\newcommand*\data{\url{https://huggingface.co/OpenAssistant/oasst1}}
\newcommand*\hforg{\url{https://huggingface.co/OpenAssistant}}

\usepackage[toc,page,header]{appendix}
\usepackage{minitoc}

\title{OpenAssistant Conversations - Democratizing Large Language Model Alignment}

\author{%
  Andreas K\"opf\thanks{These authors contributed equally to this work.} \\
  \texttt{andreas.koepf@provisio.com} \\
  \And
  Yannic Kilcher\footnotemark[1] \\
  \texttt{yannic@ykilcher.com} \\
  \AND
  Dimitri von R\"utte \\
  \And
  Sotiris Anagnostidis \\
  \And 
  Zhi-Rui Tam
  \And
  Keith Stevens \\
  \And
  Abdullah Barhoum \\
  \And
  Nguyen Minh Duc \\
  \And
  Oliver Stanley \\
  \And
  Richárd Nagyfi \\
  \And
  Shahul ES \\
  \And
  Sameer Suri \\
  \And
  David Glushkov \\
  \And
  Arnav Dantuluri \\
  \And
  Andrew Maguire \\
  \And
  Christoph Schuhmann \\
  \And
  Huu Nguyen \\
  \AND
  Alexander Mattick \\
  \texttt{alexander.mattick@googlemail.com}
}

\begin{document}

\doparttoc % Tell to minitoc to generate a toc for the parts
\faketableofcontents % Run a fake tableofcontents command for the partocs

\maketitle

\begin{abstract}
Aligning large language models (LLMs) with human preferences has proven to drastically improve usability and has driven rapid adoption as demonstrated by ChatGPT.
Alignment techniques such as supervised fine-tuning (\textit{SFT}) and  reinforcement learning from human feedback (\textit{RLHF}) greatly reduce the required skill and domain knowledge to effectively harness the capabilities of LLMs, increasing their accessibility and utility across various domains.
However, state-of-the-art alignment techniques like \textit{RLHF} rely on high-quality human feedback data, which is expensive to create and often remains proprietary.
In an effort to democratize research on large-scale alignment, we release OpenAssistant Conversations, a human-generated, human-annotated assistant-style conversation corpus consisting of 161,443 messages in 35 different languages, annotated with 461,292 quality ratings, resulting in over 10,000 complete and fully annotated conversation trees.
The corpus is a product of a worldwide crowd-sourcing effort involving over 13,500 volunteers.
Models trained on OpenAssistant Conversations show consistent improvements on standard benchmarks over respective base models.
% A preference study revealed that OpenAssistant replies are comparably preferred to GPT-3.5-turbo (ChatGPT) with a relative winrate of 48.3\% vs. 51.7\% respectively.
We release our code\footnote{\git} and data\footnote{\data} under a fully permissive licence.
\\
\\
A list of contributors who have chosen to be acknowledged by name can be found at \oacontributors.
\end{abstract}

% \vspace{-2mm}
\newpage
\section{Introduction}
\label{sec:introduction}
\vspace{-1mm}

{
Artificial intelligence (AI), particularly in the field of natural language processing, has witnessed rapid progress in recent years. Major advancements are primarily driven by a straightforward formula: take a Transformer~\cite{vaswani2017attention}-based architecture, increase the parameter count by enlarging depth and width, increase the size of the training corpus, and increase the scale of training compute. Although models have for some time exhibited an extraordinary ability to fit the training data and generalize based on their trained objective~\cite{touvron2023llama, biderman2023pythia}, their adoption among the general public has until recently been slow. This can be mainly attributed to misalignment between model predictions and final intended usage. 

The alignment of AI systems to human values, intentions, and preferences is a vital and intricate challenge within the AI research domain.
This refers to the process of ensuring that AI systems can not only successfully optimize the provided surrogate training objectives, but also that their predictions are in line with their intended purpose and adhere to ethical and safety standards provided by humans~\cite{Gabriel2020,wang2023aligning}.
One possible solution is assistant-style fine-tuning of language models that has recently emerged as a promising approach to making large language models more in line with human preferences by generating more desirable outputs based on explicitly collected human preference data~\cite{bai2022training, pmlr-v162-ethayarajh22a, thoppilan2022lamda, hilton2021webgpt,menick2022teaching,ziegler2020finetuning} and thus making them more useful.

A notable instance of such an assistant-style model is ChatGPT, which has gained unprecedented user growth due to remarkable capabilities demonstrated in a wide range of fields, but also ease-of-use for the end user~\cite{Koonchanok2023TrackingPA}. Aligning the model’s predictions is in this case accomplished by introducing human-generated examples of intended usage and using reinforcement learning from human feedback~\cite{ouyang2022training, leandro_von_werra_2023_7790115}.
In \textit{RLHF}, the human acts as a teacher and provides feedback in the form of rewards or penalties. In more detail, Ouyang et al.~\cite{ouyang2022training} proposed a three stage procedure to align language models:
    First, collect human-generated demonstrations of desired behaviour and train a supervised fine-tuned (\textit{SFT}) model.
    Second, train a reward model (RM) on human-annotated rankings for different model outputs.
    Third, use the RM as a reward function and fine-tune the \textit{SFT} model to maximize the reward generated by its responses. This is achieved using the PPO algorithm~\cite{schulman2017proximal}.

It becomes apparent that the benefits of all the aforementioned stages are predominantly dependent on the quality of the data used~\cite{DataQuality}. Despite this, availability of large-scale human feedback datasets for the open research community remains scarce.
Most openly accessible datasets are comprised of synthetic data of instructions automatically generated by querying language models~\cite{wang2022self, alpaca, anandgpt4all, peng2023instruction}. Unfortunately, these datasets are limited with respect to their complexity, creativity and quality, as they rely on a pre-specified list of possible instruction types. 
Other datasets, such as Vicuna~\cite{vicuna}, use human-generated instructions, but still rely on langauge models to produce the respective responses.
% This in turn, limits the overall performance of models trained on them that are .\todo{reference: hard to find since no such dataset exists yet ;). We can link the more general "good data good" papers, like \cite{DataQuality} }.
Without sufficiently broad and high quality data, even models with substantial size and pre-training would be inadequate for building capable, helpful, and harmless AI assistants.

Research in this area has predominantly been confined to a select few research labs with access to the required resources to engage in large-scale training and data collection.
This monopolization of access to quality data undermines the potential for inclusive and diverse research endeavours, particularly in relation to alignment challenges, which arguably constitute some of the most crucial research areas of our time.
In an effort to democratize research on aligning large language models, we introduce and release the OpenAssistant Conversations dataset.
This dataset is the culmination of an extensive open- and crowd-sourcing initiative, and its release to the research community seeks to promote more inclusive research in this highly-influential domain.
We provide a comprehensive analysis of the dataset, assessing ethical implications and safety considerations. We also fine-tune and release several assistant and preference models to further advance open access and research in this area.
% By publicly releasing both the data and models under an open license, we aim to guarantee accessibility and auditability by the wider research community.
This transparency allows for iterative improvements on the released artifacts, fostering a more collaborative and inclusive research environment.
%Our belief is that our work makes a noteworthy contribution towards creating a research landscape that is more inclusive and democratized, thereby providing opportunities to researchers from diverse backgrounds.
% It is our hope that the OpenAssistant Conversations dataset and the associated fine-tuned models will serve as a resources for the AI research community in their pursuit of developing more aligned, useful, and responsible AI systems. \todo{here we repeat the same point like many times. probably better weave in some more big-companies-bad, centralization, license shenanigans and litigation}
%We hope the resulting dataset will be an important resource for researchers studying natural language processing and machine learning, as it allows for the development and testing of new algorithms and models for conversational AI.
By providing such a large and diverse dataset, OpenAssistant Conversations opens up new avenues of research in the field, enabling researchers to explore the complexities of human language and interactions in ways that were not possible before~\cite{clark1991grounding}.
In the following sections, we delve into the intricacies of the OpenAssistant Conversations dataset and discuss its implications for the alignment of large language models and for society at large.
}

% {\red
% - Dedicated comparison with existing datasets?
% }
% \vspace{-2mm}
\newpage
\section{Data Format}
\vspace{-1mm}
The proposed dataset consists of a list of conversations.
The basic data structure is a \emph{Conversation Tree (CT)}, with nodes representing written messages in a conversation.
A CT's root node represents an initial prompt, given by the prompter.
To avoid confusion, we call the roles of the conversation \emph{prompter} and \emph{assistant}.
This allows us to reserve the term \emph{user} for the human contributors.
Although our efforts focus largely on human contributions, both the prompter and assistant roles can, in principle, be fulfilled by either a human user or a machine.
Every tree node is labelled by its role, and can have multiple children of the opposite role, each of which represents a separate next step in the conversation.
A path from the root to any node in the tree (including to itself) is called a \emph{thread}, and it represents a valid conversation with the prompter and the assistant taking turns.
Tree nodes are annotated with additional data such as user-provided labels and metadata, such as collection timestamp and indicated language.
Each \emph{assistant} node further has a rank associated which orders it compared to replies of the parent prompt, according to user preferences.

\begin{figure}[!htb]
    \centering
    \includegraphics[width=0.8\textwidth]{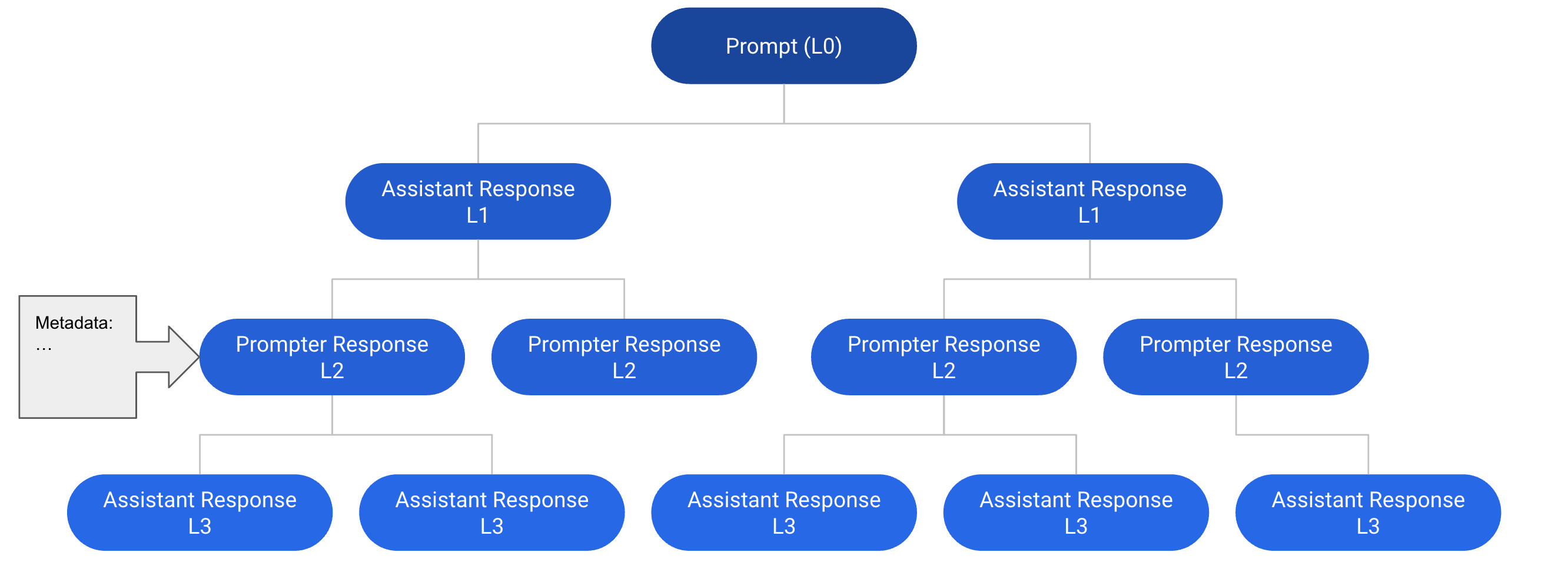}
    \caption{An example CT of depth 4 containing 12 messages. Any path from the root prompt to a node is a valid thread.}
    \label{fig:conversation_tree}
\end{figure}

\vspace{-1.5mm}
\section{Data Collection}
\label{sec:collection}
\vspace{-1mm}

The OpenAssistant Conversations dataset is a comprehensive collection of conversational data that was obtained through a crowd-sourcing effort involving more than 13,000 volunteers. The data was collected using a web-app interface, dividing the collection of each tree into five separate steps:~\emph{prompting},~\emph{labelling prompts},~\emph{adding reply messages as prompter or assistant},~\emph{labelling replies}, and~\emph{ranking assistant replies}. Users were fully informed about contributing to a public dataset. The dataset was curated with content moderation and spam filtering as key components of the annotation pipeline, ensuring high quality and safety standards.

Volunteers completed over 625,000 tasks in total, resulting in the collection of over 10,000 fully annotated and filtered Conversation Trees. 
Example User Interface (UI) displays of the data collection platform can be found in Appendix~\ref{app:webapp}, and current collection parameter settings, such as the number of collected replies to any parent message, can be found in Appendix~\ref{app:parameters}. In the following sections, we provide more details regarding the various aspects of the data collection pipeline.

\vspace{-1mm}
\subsection{Single-Step Collection}\label{sec:single-state-collection}
\vspace{-1mm}

Data collection is structured to be both efficient and effective by breaking the work down into single units and advancing multiple conversation trees one step at a time. This approach minimizes data loss due to user attrition and ensures that every unit of work is captured for utilization. The users are presented with a range of task types, either by choice or through random sampling (weighted according to current requirements). The task types include creating prompts, replying as an assistant, replying as a prompter, labeling prompts or replies, and ranking prompter or assistant replies.

\vspace{-2mm}
\paragraph{Create a prompt.} Users write an initial prompt that forms the root of a new conversation tree. As this task is particularly popular, a lottery system manages the selection of new prompts, with only a fixed number of prompts being chosen for continuation at any given moment. This method serves to regulate the influx of new prompts and maintain a balanced distribution of tasks.

\vspace{-2mm}
\paragraph{Reply as assistant.}Replying as an assistant is a more labor-intensive task that necessitates users to carefully consider their responses and often engage in external research to provide a helpful and relevant answer to the prompter's request. This task type, despite its demanding nature, has been reported to be the most enjoyable by many users due to the diverse array of topics covered. To account for the increased effort required for this task, a reward system has been implemented to incentivize users to participate. See Figure~\ref{fig:web_reply_assistant} for a UI preview.

\vspace{-2mm}
\paragraph{Reply as prompter.}The task of replying as a prompter, on the other hand, does not impose strict quality requirements but instead emphasizes on the importance of diversity to accommodate various use-cases. Examples of prompter replies may include asking for clarification, modifying the original intent, posing a follow-up question, or changing the direction of the conversation altogether.

\vspace{-2mm}
\paragraph{Label a prompt or reply.}Users are presented with a message from the database along with the preceding conversation thread (if available) and are asked to categorize the message according to three dimensions: spam detection, guideline adherence, and quality. For spam detection, users assess whether the message is unsuitable for inclusion in the dataset, for instances of obvious spam or trolling. Messages flagged as spam by multiple users are automatically removed from the dataset.

Guideline adherence is evaluated through a set of labels that determines whether the contribution aligns with the established guidelines (see Figure~\ref{fig:web_labelling}). These labels encompass the message being in a language other than the specified one, containing personally identifiable information, hate speech, sexual content, or being deemed inappropriate. Messages labelled in this manner are subsequently reviewed by human moderators.

Quality labels require users to rate the message on a five-point \textit{Likert} scale across dimensions such as quality, creativity, humorousness, politeness, and harmlessness. These labels are stored for later analysis and application. Notably, users can voluntarily assign these labels (as well as spam \& guideline adherence labels) to any message within the system, even as part of another task, as an additional contribution.

\vspace{-2mm}
\paragraph{Rank assistant replies.}Users are presented with two or more responses to the same parent message and asked to rank them in order of preference. This allows for a comparative analysis of the various responses and helps in identifying the most effective and engaging replies (Figure~\ref{fig:web_rank_assistant_replies}).

In summary, this data collection methodology effectively divides work into single units, minimizes data loss due to user attrition, and captures valuable information for future analysis and application. By offering users a diverse range of task types, the study encourages active participation and ensures the collection of rich and varied data for a comprehensive understanding of the subject.

\vspace{-1mm}
\subsection{Message Tree State Machine}
\vspace{-1mm}

The tree state machine serves as a systematic approach to managing the progression of message trees throughout the data collection process. This method ensures that each tree undergoes a series of states until it reaches completion, beginning with the creation of new trees by randomly sampling from the pool of initial prompts. The various states that a message tree passes through include the \emph{initial prompt review state}, \emph{growing state}, and \emph{end state}, as well as the \emph{aborted low-grade state} for trees that are deemed unsuitable for inclusion in the dataset and the \emph{halted by moderator} state for trees that have manually been halted by a community moderator.

Upon the creation of a new tree, it enters the \emph{initial prompt review state}, where multiple users are tasked with providing labels to assess its quality and suitability.
This state plays a crucial role in identifying any potential issues with the initial prompt, and demands special attention, as the entire rest of the tree (potentially several dozens of tasks) is rooted in the initial prompt.
If, at this point, the provided labels indicate that the tree contains spam or unsuitable content, it is transitioned to the \emph{aborted low-grade state} and subsequently removed from the dataset. Conversely, if the tree passes the \emph{initial prompt review state}, it proceeds to the \emph{growing state}.

The \emph{growing state} involves the continuous issuance of tasks to users, such as providing replies, labels, and rankings, to facilitate the development and expansion of the conversation tree. This state is essential for collecting diverse and rich data, as it allows for the accumulation of multiple interactions and the exploration of various conversation paths, given the same initial prompt. The \emph{growing state} continues until the tree reaches its \emph{end state}, which is defined by a maximum number of messages or other predetermined criteria.

Parameters within the data collection platform govern the behaviour of the tree state machine, such as the average number of messages collected for each parent message or the maximum tree depth. These parameters enable researchers to fine-tune the data collection process according to their specific research goals and requirements, ensuring a more targeted and efficient approach to gathering data.
Parameters varied during the collection of the dataset.
Current settings can be found in Appendix~\ref{app:parameters}.

In summary, the tree state machine is a structured and systematic method for managing the progression of message trees during the data collection process. By guiding each tree through a series of states, from initial prompt review to growing and reaching its \emph{end state}, the tree state machine ensures the collection of high-quality, diverse, and relevant data. Additionally, the inclusion of platform parameters allows for the customization of the data collection process to align with specific research objectives, further enhancing the effectiveness and utility of this approach.

\vspace{-1mm}
\subsection{Contributor Guidelines}
\vspace{-1mm}

To achieve a high degree of quality and consistency across a wide range of contributors, we issue clear and detailed guidelines.
A full copy of these guidelines at the present time can be found in Appendix~\ref{app:contributor_guidelines}.
Our guidelines follow three main goals: First, clarify the meanings, scales, and criteria for assigning labels and rankings during the labelling and ranking tasks. Second, make assistant responses be polite, helpful, concise, friendly, and safety-aware and third, instruct prompts and prompter replies to explore a diverse and challenging set of inputs to the assistant role.

The guidelines establish a framework for safely interacting with an automated assistant by drawing inspiration from the concept of \emph{informed consent}.
Rather than categorically denying large parts of request categories, we aim to provide the prompter with useful feedback, for example drawing special awareness to dangerous activities, elaborating on weaknesses of automated assistants, such as hallucinations, and discouraging and denying requests asking for illegal or highly inappropriate content.
In our validation experiments in training assistant models based on OpenAssistant Conversations, we observe a high degree of consistency of the trained models' outputs with our given guidelines.

Although guideline adherence is already high in our models after training, our approach is completely compatible with deploying additional safety measures during inference, such as secondary models to filter or modify ill-suited user input.

\vspace{-1mm}
\subsection{Quality Control \& Content Moderation}
\vspace{-1mm}

We take a multi-pronged approach to quality assurance, with the main pillars being a system of reward points \& leaderboards, and manual review of flagged content by human moderators.
This both maximizes the quality of contributions, while effectively utilizing the limited time of the volunteer moderators. In an effort to demonstrate progress and achievement to users, and to encourage high-quality contributions, our system allocates points for the completion of tasks. These points contribute to various leaderboards, including daily, weekly, monthly, and all-time rankings. A level system also exists, wherein higher point accumulation results in elevated levels, reflecting veteran status and engagement. In the future, this system could potentially be developed further to facilitate preferential access to more engaging tasks or similar perks.

The distribution of points is contingent upon task type, as certain tasks require greater effort, such as the \emph{reply as assistant} task (compared to the \emph{create a prompt} task). A significant portion of points is deferred and reliant on interactions with other users. For instance, a user's assistant reply may gather many additional points if it is subsequently deemed non-spam and highly ranked by other users. Inversely, points may be reduced or lost for answers that are labeled as spam or down-voted by consensus of other users.

Within the moderator section of the website, an alternative leaderboard, designated the \emph{Trollboard}, is exhibited. This leaderboard assesses users based on an aggregate of negative labels, reports, and down-votes received for their contributions. This approach enables human moderators to proactively scrutinize potentially misbehaving users in a comprehensive manner. The Trollboard has proven to be an effective tool in addressing the numerical disparity between users and moderators, maximizing the collective efforts of contributors to identify undesirable contributions.

Users further have the option to report messages to moderators for manual review, either via the platform, or directly via communication on a community chat server. Moderators have the ability to delete individual messages, or all messages of a given user, at their own discretion. Deleted messages are retained, but marked as deleted and are not exported for training.

\newpage
\section{Dataset Composition}
\vspace{-1mm}

% We release several variants of the OpenAssistant Conversations dataset representing various levels of filtering. 
The full dataset consists of 161,443 messages (91,829 prompter and 69,614 assistant messages) distributed across 66,497 conversation trees, in 35 different languages, annotated with 461,292 quality ratings. This includes 8,576 synthetic messages, leaving 152,867 human-submitted messages. Of the 66,497 total conversation trees, we consider 10,968 complete, meaning the full number of messages has been collected and the moderation process for these trees has been concluded. 52,159 incomplete trees are in the prompt lottery state, meaning they only consist of a single initial prompt. The completed trees contain 92,365 messages.

The set of categories for which \textit{Likert}-scale human labels are collected is Creativity, Quality, Humor, Helpfulness, Violence, and Rudeness. The set of categories for which binary human labels are collected is Language Mismatch, Not Appropriate, Personally Identifiable Information, Hate Speech, and Sexual Content. We additionally release the rank of each assistant message compared to other assistant messages submitted for the same prompt, computed from the preference rankings of several human annotators.
To merge the rankings of multiple (possibly conflicting) annotators, we use a variant of Tideman's method, described in Appendix~\ref{app:ranking}.

%Of the 161,443 total messages, 69,614 are assistant replies and 91,829 are user prompts. Related to this, 52,159 conversation trees consist of only a single initial user prompt which has not yet received any assistant replies.

\iffalse
{\red

- distribution of quality labels (quality, humorous, etc.)

- distribution of length of messages

- amount of time people spent per task type

- show topic models (-> currently word-clouds in appendix)

}
\fi

\begin{figure}[H]
    \centering
    \includegraphics[width=0.39\linewidth]{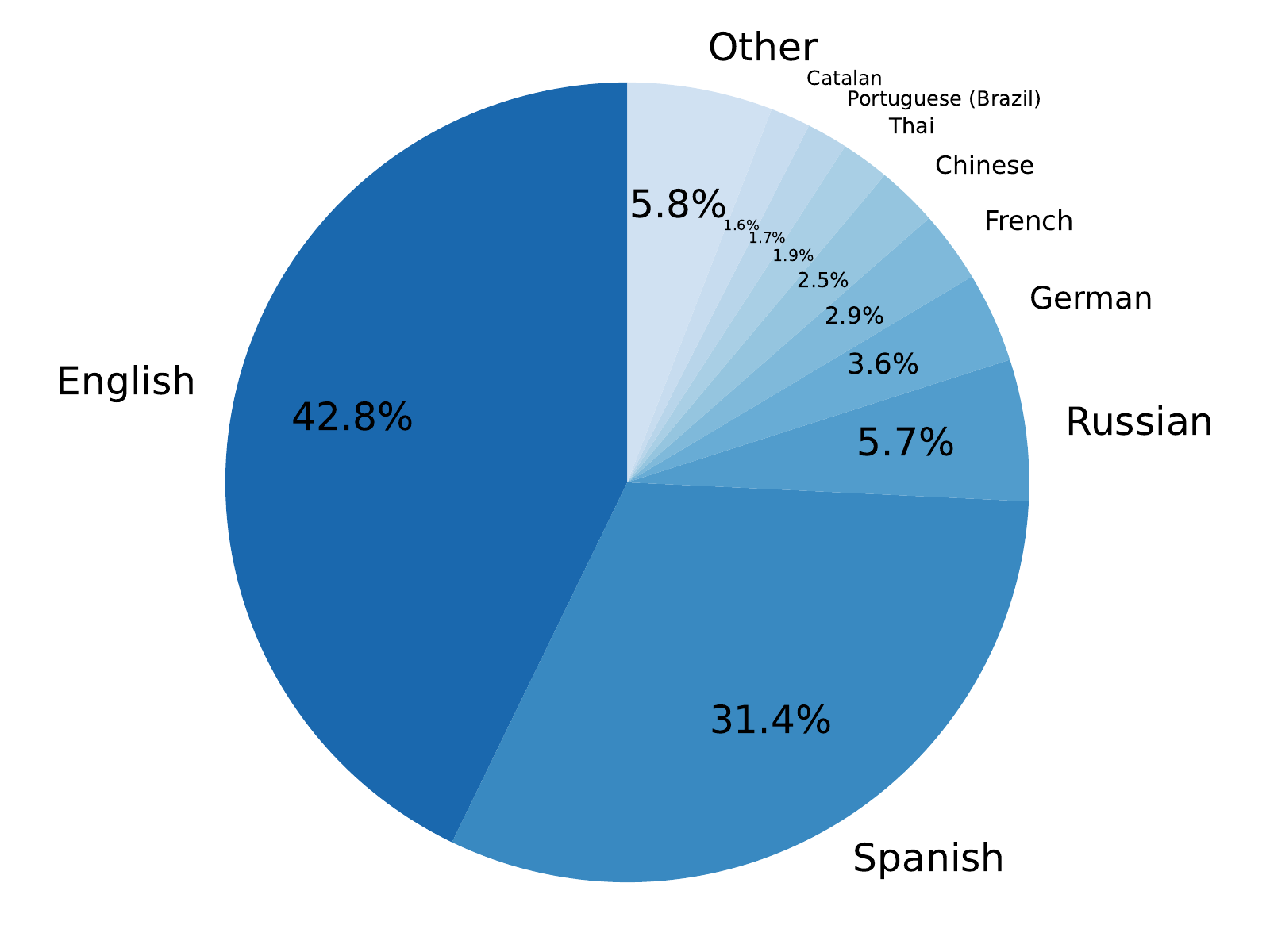}
    \includegraphics[width=0.59\linewidth]{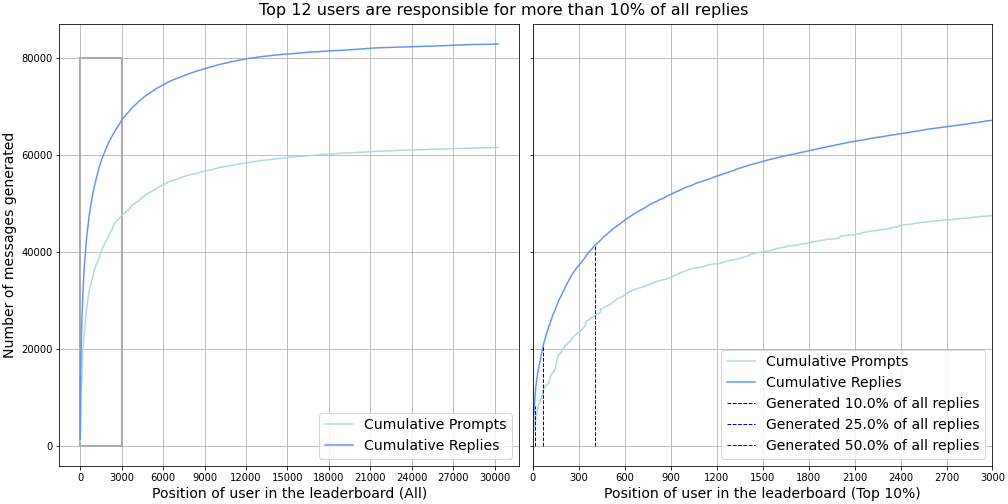}
    \caption{\textbf{Left:} Relative share of the most frequent languages in the dataset. \textbf{Right:} Distribution of contributions among users.}
    \label{fig:language_pie}
\end{figure}

The dataset is dominated by English and Spanish messages as illustrated in Figure~\ref{fig:language_pie} (left). The prominence of English is expected as a result of the community around OpenAssistant originating in the English-speaking open-source machine learning community. The high quantity of Spanish messages can be attributed to the publicity given to OpenAssistant by prominent figures in the Spanish machine learning community.
Figure~\ref{fig:language_pie} (right) illustrates how a small number of power users contributed a significant proportion of the dataset. This must be taken into account when considering possible biases in the data. Although significant effort went into preventing responses directly copied from other sources, it is possible that some users utilised automated techniques to enter data.

% \begin{wrapfigure}{r}{0.5\textwidth}
%     \centering
    % \includegraphics[width=0.49\linewidth]{images/survey/top.png}
%     \caption{Distribution of Messages}
%     \label{fig:topdist}
% \end{wrapfigure}

We release the dataset on the Hugging Face Hub\footnote{\data} in several variants: One variant containing all collected initial prompts, one variant containing all trees that are considered completed, one variant containing all trees (completed and in-progress), and one variant containing messages filtered out as spam.
For most purposes, such as instruction-tuning of language models, we recommend using the variant containing all completed trees, and include non-completed trees if more data is required.

\vspace{-1mm}
\section{Contributor Demographics and Satisfaction}
\label{sec:demographics}
\vspace{-1mm}

{
To gain a deeper understanding of the contributors' demographics, a Google Form survey was sent out as an announcement on the project's Discord channel.
The survey consists of 3 parts with questions on demographic information, personal motivation and user satisfaction. 
%Since prompts are received from all over the world in multiple languages, we wanted to make sure that demographic questions, such as levels of completed education~\cite{hoffmeyer2005measure} are in fact relatable to everyone.
%We have omitted questions on ethnicity and nationality. Instead, we asked questions about English proficiency, country of origin, and the language for which most contributions were made.
At the time of the release of this paper, a total of 270 survey answers have been collected.
Results can be seen in Figures~\ref{fig:demography},\ref{fig:familiarityAI},\ref{fig:Motivation},\ref{fig:UseCases}, and Tables~\ref{tab:LikertSatisfaction} and~\ref{tab:contributePrevious} (all in the Appendix).
%243 people, the overwhelming majority of the respondents, were male and only 11 of the respondents female. Only 6 of our respondents self-identified as non-binary / other, and 10 preferred not to answer. 
%Respondent answers indicate , the respondents do differ in their level of education and motivation (figure~\ref{fig:Motivation}) for contribution. They understand and use artificial intelligence at different levels (figure~\ref{fig:familiarityAI}) and have different use cases for the technology (figure~\ref{fig:UseCases}).    
%People were in general very happy to have contributed to the project, with  95.55\% either agreeing or strongly agreeing with the statement "Overall, I’m glad I have contributed to OpenAssistant." (figure~\ref{tab:LikertSatisfaction}) For about 40\% (figure~\ref{tab:contributePrevious}), this has been their very first time contributing to a community project.
Respondent answers indicate a homogeneity towards identifying as male, but strong diversity in the reported levels of education, stated motivations for contribution, proficiency in artificial intelligence, and use cases for the technology.
Over 95\% of respondents either agreed or strongly agreed with the statement "Overall, I’m glad I have contributed to OpenAssistant." About 40\% reported this being their first time contributing to an open-source project. We note that the method of recruiting via Discord is biased towards users who are present on the platform and have been active around the time of the announcement. 
%Therefore, we intend to send out e-mails to registered users in the future. Fluency in English can also affect the willingness to participate in the survey. Translations of the survey are planned in languages with higher representations, such as Spanish, to ensure that responses from monolingual users are not missed. 
Active contributors are also expected to be more likely to respond.
}
\vspace{-1mm}
\section{Experimental Validation}
\vspace{-1mm}
%In an effort to demonstrate the effectiveness of the OpenAssistant Conversations dataset and to validate the aforementioned collection pipeline, we pursue to reproduce the results of ChatGPT~\cite{ouyang2022training}.

{\red
\iffalse
- we train and release a suite of fine-tuned language models based on pythia \cite{biderman2023pythia} (permissive open-source license) as well as llama \cite{touvron2023llama} (bespoke non-commercial license).

- the model release includes instruction-tuned pythia-12B, llama-13B as well as llama-30B, our largest model to date.

- we evaluate our pythia-12B model in detail, as it is the biggest model with an unrestrictive license. we analyze its performance by conducting a user preference study where we compare its performance to OpenAI's \texttt{gpt-3.5-turbo} model.
As of the time of writing, the study has received 348 submissions, amounting to 7042 comparisons.
Ignoring ties (which occur 16.4\% of the time), our \emph{Pythia} model has a winrate of 48.3\% (95\% confidence interval of $\pm$ 1.28\%, $N=5,889$) against GPT, suggesting that its answers are 93.5\% as preferable as GPT
\fi
}

\vspace{-1mm}
\subsection{Instruction Tuning \& Preference Modeling}
\label{sec:instruction_tuning}
\vspace{-1mm}
We focus on the development and evaluation of fine-tuned language models based on Pythia~\cite{biderman2023pythia}, LLaMA~\cite{touvron2023llama}, and Falcon~\cite{falcon40b}. Pythia and Falcon are state-of-the-art language models with permissive open-source licenses, while LLaMA is a powerful language model with a bespoke non-commercial license. Specifically we train supervised fine-tuned (SFT) models, reward models (RM~\cite{ouyang2022training},\cite{askell2021general}), and, using the trained reward models, reinforcement-learned models (RLHF)\footnote{All trained models are released at \hforg}.

\begin{table}[h]
    \centering
    \begin{tabular}{lcccc}
    \toprule
    Model & \textbf{LMEH} & \textbf{VEL} & \textbf{OAIE} & \textbf{HE} \\
    \midrule
    gpt-3.5-turbo (ChatGPT) & & 1110 & 0.87 & 0.72 \\
    EleutherAI/pythia-12b & 60.33 \\
    OpenAssistant/pythia-12b-sft-v8-7k-steps & 60.28 & 997 & 0.10 & 0.10 \\
    tiiuae/falcon-40b & 72.29 \\
    OpenAssistant/falcon-40b-sft-top1-560 & 74.04 & 1192 & 0.26 & 0.09 \\
    OpenAssistant/falcon-40b-sft-mix-1226 & 74.40 & 1053 & 0.44 & 0.13 \\
    huggyllama/llama-65b & 67.24 \\
    OpenAssistant/oasst-sft-7e3-llama-30b & 68.03 & 979 & 0.52 & 0.20 \\
    OpenAssistant/oasst-rlhf-3-llama-30b-5k-steps & 68.51 & 1068 & 0.51 & 0.15 \\
    \bottomrule
    \end{tabular}
    \caption{Comparison of model evaluation scores on different LLM benchmarks:
    \textbf{LMEH:} lm-evaluation-harness \cite{eval-harness} (average scores, see online leaderboard for more details)
    \textbf{VEL:} Vicuna Elo Rank \cite{vicuna}
    \textbf{OAIE:} OpenAI Evals \cite{openai2023gpt4}
    \textbf{HE:} HumanEval \cite{chen2021codex}
    (for all benchmarks, higher is better).
    We have chosen to leave the Hugging Face Hub identifiers as the model names for identifiability.
    }
    % lm-evaluation-harness	& Vicuna Elo Rank & OpenAI Evals & HumanEval \\
    \label{tab:llm_evals}
\end{table}

Table~\ref{tab:llm_evals} shows evaluation scores of a selection of baseline and trained models on a set of standard benchmark datasets. Evaluations were performed externally using FastEval and evaluation results are hosted on a leaderboard, which is continually being updated\footnote{\url{https://github.com/FastEval/FastEval}, \url{https://tju01.github.io/ilm-eval/}}.
LMEH refers to the average performance on a set of widely-used NLU tasks consisting of BoolQ, PIQA, HellaSwag, WinoGrande, ARC-e, ARC-c and OBQA\footnote{For readability, Table~\ref{tab:llm_evals} contains aggregated LMEH scores. Details in the online leaderboard.}.
We omit instruction-centric experiments (VEL, OAIE, HE) for the base models, as these benchmarks are unsuitable for non-instruction-tuned models.
The results show that models using OpenAssistant Conversations are consistently outperforming the corresponding baseline models (in the case of LLaMA even a larger baseline model). RLHF outperforms SFT in some benchmarks, but not in others. 
For Falcon-based models, \textit{sft-top1} is trained only on top-ranked conversation threads, whereas \textit{sft-mix} mixes OpenAssistant Conversations with other instruction datasets (details in Appendix~\ref{sec:training_config}).
The varied evaluation scores demonstrate that by combining different data sources, the nature of the resulting model can be readily influenced.
Ranks across benchmarks are not consistent, which could indicate the unsuitability of automatic evaluations for language models, or could indicate that different models and datasets lead to different capabilities.
The results also show that while open-source models are close to matching ChatGPT in some benchmarks, others still show large performance gaps.
%We present more results in the Appendix.
Anecdotally, users report OpenAssistant models to be less robotic and more human-sounding than commercial models and report generations to have high quality and diversity in domains such as creative writing, conversational messaging, and drafting social media posts.

\iffalse
{
To evaluate the performance of Pythia-12B, we conducted a user preference study comparing its output to that of OpenAI's \texttt{gpt-3.5-turbo} model. As of the time of writing, this study has garnered 348 submissions, amounting to a total of 7042 comparisons. After excluding ties, which account for 16.4\% of the total comparisons, we found that Pythia-12B has a win rate of 48.3\% (95\% confidence interval of $\pm$ 1.28\%, $N=5,889$) against \texttt{gpt-3.5-turbo}. This result implies that the answers generated by Pythia-12B are 93.5\% as preferable as those produced by \texttt{gpt-3.5-turbo}, indicating that our fine-tuned Pythia model is a strong competitor in the realm of large-scale language models.
For more details on the user preference study, we refer to Appendix~\ref{sec:user-preference-study}.
}
\fi
% \vspace{-2mm}
% \subsection{Preference Modelling}
% \vspace{-2mm}

{\red
\iffalse
- additionally we also release trained reward models based on pythia-1.4B and pythia-12B which are trained on OpenAssistant data using the reward model training regime from \cite{ouyang2022training}.

- ultimately, we're planning on releasing llama-30B models trained on RLHF, but development in this area is ongoing.
\fi
}

\vspace{-1mm}
\newpage
\subsection{Spam and Toxicity}
\label{sec:spam_and_toxicity}
\vspace{-1mm}

To understand the concordance between human and automated toxicity detection, we employ toxicity detection methods based on Detoxify \cite{Detoxify} to obtain automated ratings for six distinct categories, classifying whether a message is toxic, obscene, threatening, insulting, attacking a certain identity or sexually explicit. We limit our analysis to those languages that are supported by the toxicity detection method, covering English, Spanish, Russian, French, and Italian. These languages represent the majority of OASST1 messages (over 83\%).

\begin{wrapfigure}{r}{0.5\textwidth}
    \centering
    \includegraphics[width=.9\linewidth]{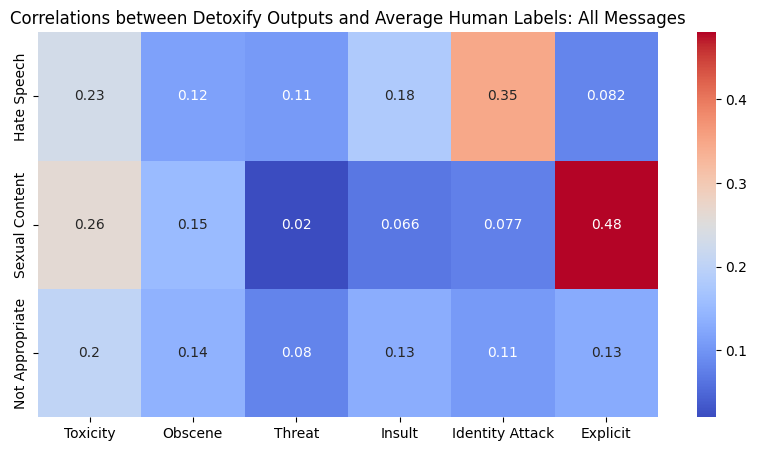}
    \caption{Correlation between human labels and Detoxify outputs for all messages in Detoxify-supported languages.}
    \label{fig:corr_detoxify_all}
\end{wrapfigure}

{

Using automated toxicity ratings, we are able to systematically assess the correlation between these ratings and human-assigned toxicity labels (hate speech, not appropriate, and sexual content).
Based on a sample of 115,153 messages, we compute the correlation between automatic and human-annotated toxicity labels, which is visualized in Figure~\ref{fig:corr_detoxify_all}.
We see a correlation between human and automatic labels in at least one element of each row and column of the correlation matrix, suggesting agreement between human annotators and off-the-shelf toxicity detection models.
The results serve to both validate the capabilities and show limitations of AI-driven toxicity detection in comparison to human judgement and may inform future work in this area.

In addition to analysing the correlation between human-assigned toxicity labels and automated ratings, we extend the application of the Detoxify model to assess the efficacy of the moderation process for the same languages described earlier.
To facilitate this analysis, we define two categories of messages: \emph{deleted} messages, which encompass those that either failed to pass the community moderation process or were subsequently manually removed by moderators, and \emph{retained} messages, which successfully made it through to the dataset. In order to provide a comprehensive evaluation of the moderation process, we calculated average values for each of the six Detoxify categories for both \emph{deleted} and \emph{retained} messages.
The values obtained for this analysis are based on a sample of 74,781 messages. We excluded messages in trees that were incomplete at the time of export, as these messages may be subject to removal by the moderation process.

Our analysis, presented in Table ~\ref{tab:toxicity_deleted_retained} shows that the values for all six toxicity categories are markedly higher for \emph{deleted} messages compared to \emph{retained} messages. This significant difference demonstrates the effectiveness of the moderation processes in place, as messages removed from the dataset are on average rated as significantly more toxic by the Detoxify model than messages allowed to remain in the dataset.

\begin{table}[h]
    \centering
    \begin{tabular}{lccccccc}
    \toprule
     & Toxicity & Obscene & Threat & Insult & Identity Attack & Explicit & N \\
    State &  &  &  &  &  &  &  \\
    \midrule
    Deleted & 4.625\% & 1.965\% & 0.411\% & 2.085\% & 0.651\% & 1.39\% & 3422 \\
    Retained & 0.988\% & 0.574\% & 0.102\% & 0.715\% & 0.121\% & 0.177\% & 71359 \\
    \bottomrule
    \end{tabular}
    \caption{Detoxify outputs across six categories of toxicity, comparing \emph{deleted} and \emph{retained} messages.}
    \label{tab:toxicity_deleted_retained}
\end{table}

While \emph{deleted} messages are rated as more toxic than \emph{retained} messages by the Detoxify model across all categories, the average toxicity values for these messages are still small. This implies toxicity ratings from models like Detoxify alone are not sufficient to determine when messages are unsuitable for inclusion in the dataset. Reasons for deleting non-toxic messages may include a lack of factual accuracy, or poor grammar. Additionally, messages which are children of deleted messages must themselves be deleted even if they appear to be acceptable in isolation.

}

{\red
\iffalse

- run the data through some spam \& toxicity model and see overlap with human labels

- investigate disagreements between humans \& models, hopefully humans are mostly correct %\todo{richard: there are multiple mentions about human disagreement, we should back it up with examples}
%\todo{richard: also, there are cases where the 2 best answers are ranked really similarly, which adds noise to the tree seleciton}
\fi
}

\newpage
\section{Limitations}
\label{sec:limitations}
\vspace{-1mm}

{\red
\iffalse
- subjective and cultural biases: the dataset is bound to reflect any biases of the annotators and due to the open nature of the project, it's difficult to control for this. the demographics of the annotators are very diverse (all over the world, anyone can contribute) and homogeneous (similar interests, similar age, predominantly male) at the same time, with 89.1\% of users identifying as male with a median age of 26.

- although the number of contributors is large, different users contributed at different rates, with the more engaged users providing more contributions than others. this means that the dataset will represent the values and interests of the most active users more. 

- while we do have measures to detect harmful messages, our system is not perfect and it's possible that the dataset contains unsafe content. we believe the open nature of the project allows for the data filtering to be done openly and transparently and converge on the highest standard possible.

- as such, we advocate using the data for academic research only and strongly advise careful investigation of safety and bias in downstream models.

- Aligning LLMs is important and we hope that our contributions can help advance the field of AI alignment.
However, current alignment techniques are not perfect and can even exacerbate certain biases \cite{glaese2022improving}.
We stress that the released models are for research purposes only as they can exhibit unsafe behavior and are likely vulnerable to prompt injection attacks.
\fi
}

\textbf{Reward model data collection.}
InstructGPT~\cite{ouyang2022training} trained reward models on ranking data of messages generated by their initial SFT model, while our reward models are trained using ranking data of human-generated messages.
We chose to do so because we were already collecting this ranking data as part of our general efforts, for use in quality control, spam filtering, and dataset sub-sampling.
While subjectively, many users report our RLHF models to follow instructions more closely (also compare Section~\ref{sec:instruction_tuning}), the models do not deliver the same uniform and significant improvements over SFT models as reported in \cite{ouyang2022training}.
We hypothesize that the difference in data collection for the reward model could at least partially explain this gap.
We plan to collect ranking data based on our own SFT models in the future to verify these assumptions.
Further research is necessary to determine more precise criteria for collecting data useful to RLHF.

\textbf{Subjective, Cultural, and Contribution Frequency Biases.}
The open nature of our project introduces a unique set of challenges when it comes to controlling for biases within the dataset. Annotators from diverse backgrounds contribute to the dataset, with demographics that are simultaneously heterogeneous in some dimensions and homogeneous in others (see Section~\ref{sec:demographics}). Specifically, 89.1\% of the annotators identify as male, with a median age of 26. This demographic profile may inadvertently introduce biases in the dataset, as it is bound to reflect the values, perspectives, and interests of the annotators.
(We pose that some of this could be mitigated by introducing more constrained conversations, for example sampling a random Wikipedia page to determine the conversation topic.)
Further, users' participation levels differ significantly. More engaged users contribute a greater number of annotations (see Figure~\ref{fig:language_pie}), which likely leads to over-representation of their values and interests in the dataset. Consequently, the dataset may not adequately capture the diverse perspectives that a more balanced distribution of contributions could have provided.
Further research is necessary to determine the effect of uneven contributor distributions have when given a clear, general task.

\textbf{Possibility of Unsafe Content.}
While we have implemented measures to detect and remove harmful messages, our system is not infallible. It is possible that the dataset still contains unsafe content. We believe that the open nature of the project allows for data filtering to be conducted in a transparent manner, ultimately converging on the highest possible standards. Nevertheless, the potential presence of residual unsafe content in the dataset necessitates careful evaluation of any models trained on it.

Given the limitations discussed above, we advocate for the use of our models in academic research contexts only. We strongly encourage researchers to thoroughly investigate the safety and bias of any model before employing it in downstream tasks. The released models may exhibit unsafe behavior and are likely susceptible to prompt injection attacks. The alignment of LLMs is a crucial aspect of AI research, and we hope that our contributions can help advance the field of AI alignment. However, we also acknowledge that current alignment techniques are not perfect and can even exacerbate certain biases~\cite{glaese2022improving}. As such, researchers should exercise caution when using these models and be cognizant of their limitations.
Additionally, it is essential to continue refining alignment techniques and advancing the field of AI alignment in order to mitigate these limitations and develop more reliable and robust LLMs.

\vspace{-1mm}
\section{Safety and Ethical Implications}
\label{sec:safety}
\vspace{-1mm}

Large language models are prone to generating inaccurate information about people, places, or facts, commonly known as `hallucinations'~\cite{kryscinski2019evaluating, shuster2021retrieval}. LLMs can also produce toxic or hateful content and fail to follow provided constraints~\cite{henderson2018ethical}.
Additionally, these models tend to incorporate biases present in their training data, leading to unfair and discriminatory outputs~\cite{dixon2018measuring}. While methods such as \textit{RLHF} can mitigate some of these shortcomings, they may exacerbate others~\cite{perez2022discovering, glaese2022improving}.
We hope that alignment methods using OpenAssistant Conversations can fix some of these issues~\cite{ouyang2022training}, but we acknowledge that achieving full alignment is a complex and ongoing challenge. 

We recognize that sufficiently powerful language models can have a significant impact on society~\cite{weidinger2021ethical}, and therefore we believe it is essential to promote transparency in their development and deployment.
OpenAssistant Conversations is our contribution to this goal of transparency.
%In addition to keeping their weights private, many commercial models also disguise their model sizes and architectures, compute requirements, and training data, which leads to diminished auditability of the safety of the resulting models.

%Our team has put in significant effort to ensure that the community has access to an open-source high-quality dataset free of unethical or harmful responses. We believe that creating a safe and respectful environment for our users is paramount, and we encourage them to generate prompts and replies that are not only polite, but also creative and detailed. To ensure the quality of our dataset, we have established strict contributor guidelines that all users must follow. These guidelines are designed to prevent harmful content from being added to our dataset, and to encourage contributors to generate high-quality responses. Previous sections and the contributor guidelines in Appendix~\ref{app:contributor_guidelines} provide detailed information. Overall, our goal is to create a dataset that is both useful and safe for future research. We believe that it is essential to conduct alignment research at an appropriate pace relative to improving general capabilities. By releasing the OpenAssistant Conversations dataset, we hope to facilitate further research in this area.

{}

\begin{ack}

Our greatest thanks go to the many volunteer contributors, of human data, code, moderation, documentation, and community organization.
Absent of any financial incentives, this project is a stunning and unprecedented display of global cooperation of humans for the purpose of advancing and democratizing AI research.
In addition, several organizations have contributed to this project with resources:
Redmond AI provided training compute.
Stability AI and Hugging Face provided inference compute.
We thank Olivier Dehaene at Hugging Face for close collaboration and personal support.
Weights \& Biases provided their full MLOps solution to the entire team.
LAION provided legal input and acts as the website addressee.
We thank Til Jasper Ullrich for running LLM benchmark evaluations,
Luke Thomas Kaiser for running evaluations on model bias,
and Silvia Pareti for detailed feedback on the manuscript.
Agradecemos a Carlos Santana por la promoción a gran escala de la plataforma de recopilación de datos para la comunidad hispanohablante.\footnote{Translated from English by OpenAssistant, first try}

\end{ack}

%Bibliography
\bibliographystyle{unsrt}  
\bibliography{references}

\begin{thebibliography}{10}

\bibitem{vaswani2017attention}
Ashish Vaswani, Noam Shazeer, Niki Parmar, Jakob Uszkoreit, Llion Jones,
  Aidan~N Gomez, {\L}ukasz Kaiser, and Illia Polosukhin.
\newblock Attention is all you need.
\newblock {\em Advances in neural information processing systems}, 30, 2017.

\bibitem{touvron2023llama}
Hugo Touvron, Thibaut Lavril, Gautier Izacard, Xavier Martinet, Marie-Anne
  Lachaux, Timoth{\'e}e Lacroix, Baptiste Rozi{\`e}re, Naman Goyal, Eric
  Hambro, Faisal Azhar, et~al.
\newblock Llama: Open and efficient foundation language models.
\newblock {\em arXiv preprint arXiv:2302.13971}, 2023.

\bibitem{biderman2023pythia}
Stella Biderman, Hailey Schoelkopf, Quentin Anthony, Herbie Bradley, Kyle
  O'Brien, Eric Hallahan, Mohammad~Aflah Khan, Shivanshu Purohit, USVSN~Sai
  Prashanth, Edward Raff, et~al.
\newblock Pythia: A suite for analyzing large language models across training
  and scaling.
\newblock {\em arXiv preprint arXiv:2304.01373}, 2023.

\bibitem{Gabriel2020}
Iason Gabriel.
\newblock Artificial intelligence, values, and alignment.
\newblock {\em Minds and Machines}, 30(3):411--437, September 2020.

\bibitem{wang2023aligning}
Yufei Wang, Wanjun Zhong, Liangyou Li, Fei Mi, Xingshan Zeng, Wenyong Huang,
  Lifeng Shang, Xin Jiang, and Qun Liu.
\newblock Aligning large language models with human: A survey, 2023.

\bibitem{bai2022training}
Yuntao Bai, Andy Jones, Kamal Ndousse, Amanda Askell, Anna Chen, Nova DasSarma,
  Dawn Drain, Stanislav Fort, Deep Ganguli, Tom Henighan, Nicholas Joseph,
  Saurav Kadavath, Jackson Kernion, Tom Conerly, Sheer El-Showk, Nelson Elhage,
  Zac Hatfield-Dodds, Danny Hernandez, Tristan Hume, Scott Johnston, Shauna
  Kravec, Liane Lovitt, Neel Nanda, Catherine Olsson, Dario Amodei, Tom Brown,
  Jack Clark, Sam McCandlish, Chris Olah, Ben Mann, and Jared Kaplan.
\newblock Training a helpful and harmless assistant with reinforcement learning
  from human feedback, 2022.

\bibitem{pmlr-v162-ethayarajh22a}
Kawin Ethayarajh, Yejin Choi, and Swabha Swayamdipta.
\newblock Understanding dataset difficulty with $\mathcal{V}$-usable
  information.
\newblock In Kamalika Chaudhuri, Stefanie Jegelka, Le~Song, Csaba Szepesvari,
  Gang Niu, and Sivan Sabato, editors, {\em Proceedings of the 39th
  International Conference on Machine Learning}, volume 162 of {\em Proceedings
  of Machine Learning Research}, pages 5988--6008. PMLR, 17--23 Jul 2022.

\bibitem{thoppilan2022lamda}
Romal Thoppilan, Daniel~De Freitas, Jamie Hall, Noam Shazeer, Apoorv
  Kulshreshtha, Heng-Tze Cheng, Alicia Jin, Taylor Bos, Leslie Baker, Yu~Du,
  YaGuang Li, Hongrae Lee, Huaixiu~Steven Zheng, Amin Ghafouri, Marcelo
  Menegali, Yanping Huang, Maxim Krikun, Dmitry Lepikhin, James Qin, Dehao
  Chen, Yuanzhong Xu, Zhifeng Chen, Adam Roberts, Maarten Bosma, Vincent Zhao,
  Yanqi Zhou, Chung-Ching Chang, Igor Krivokon, Will Rusch, Marc Pickett,
  Pranesh Srinivasan, Laichee Man, Kathleen Meier-Hellstern, Meredith~Ringel
  Morris, Tulsee Doshi, Renelito~Delos Santos, Toju Duke, Johnny Soraker, Ben
  Zevenbergen, Vinodkumar Prabhakaran, Mark Diaz, Ben Hutchinson, Kristen
  Olson, Alejandra Molina, Erin Hoffman-John, Josh Lee, Lora Aroyo, Ravi
  Rajakumar, Alena Butryna, Matthew Lamm, Viktoriya Kuzmina, Joe Fenton, Aaron
  Cohen, Rachel Bernstein, Ray Kurzweil, Blaise Aguera-Arcas, Claire Cui,
  Marian Croak, Ed~Chi, and Quoc Le.
\newblock Lamda: Language models for dialog applications, 2022.

\bibitem{hilton2021webgpt}
Jacob Hilton, R~Nakano, S~Balaji, and John Schulman.
\newblock Webgpt: Improving the factual accuracy of language models through web
  browsing.
\newblock {\em OpenAI Blog, December}, 16, 2021.

\bibitem{menick2022teaching}
Jacob Menick, Maja Trebacz, Vladimir Mikulik, John Aslanides, Francis Song,
  Martin Chadwick, Mia Glaese, Susannah Young, Lucy Campbell-Gillingham,
  Geoffrey Irving, et~al.
\newblock Teaching language models to support answers with verified quotes.
\newblock {\em arXiv preprint arXiv:2203.11147}, 2022.

\bibitem{ziegler2020finetuning}
Daniel~M. Ziegler, Nisan Stiennon, Jeffrey Wu, Tom~B. Brown, Alec Radford,
  Dario Amodei, Paul Christiano, and Geoffrey Irving.
\newblock Fine-tuning language models from human preferences, 2020.

\bibitem{Koonchanok2023TrackingPA}
Ratanond Koonchanok, Ya-Chen Pan, and Hyeju Jang.
\newblock Tracking public attitudes toward chatgpt on twitter using sentiment
  analysis and topic modeling.
\newblock {\em ArXiv}, abs/2306.12951, 2023.

\bibitem{ouyang2022training}
Long Ouyang, Jeff Wu, Xu~Jiang, Diogo Almeida, Carroll~L. Wainwright, Pamela
  Mishkin, Chong Zhang, Sandhini Agarwal, Katarina Slama, Alex Ray, John
  Schulman, Jacob Hilton, Fraser Kelton, Luke Miller, Maddie Simens, Amanda
  Askell, Peter Welinder, Paul Christiano, Jan Leike, and Ryan Lowe.
\newblock Training language models to follow instructions with human feedback,
  2022.

\bibitem{leandro_von_werra_2023_7790115}
Leandro von Werra, Jonathan Tow, reciprocated, Shahbuland Matiana, Alex
  Havrilla, cat state, Louis Castricato, Alan, Duy~V. Phung, Ayush Thakur,
  Alexey Bukhtiyarov, aaronrmm, Fabrizio Milo, Daniel, Daniel King, Dong Shin,
  Ethan Kim, Justin Wei, Manuel Romero, Nicky Pochinkov, Omar Sanseviero,
  Reshinth Adithyan, Sherman Siu, Thomas Simonini, Vladimir Blagojevic,
  Xu~Song, Zack Witten, alexandremuzio, and crumb.
\newblock {CarperAI/trlx: v0.6.0: LLaMa (Alpaca), Benchmark Util, T5 ILQL,
  Tests}, March 2023.

\bibitem{schulman2017proximal}
John Schulman, Filip Wolski, Prafulla Dhariwal, Alec Radford, and Oleg Klimov.
\newblock Proximal policy optimization algorithms.
\newblock {\em arXiv preprint arXiv:1707.06347}, 2017.

\bibitem{DataQuality}
Lukas Budach, Moritz Feuerpfeil, Nina Ihde, Andrea Nathansen, Nele Noack,
  Hendrik Patzlaff, Felix Naumann, and Hazar Harmouch.
\newblock The effects of data quality on machine learning performance, 2022.

\bibitem{wang2022self}
Yizhong Wang, Yeganeh Kordi, Swaroop Mishra, Alisa Liu, Noah~A Smith, Daniel
  Khashabi, and Hannaneh Hajishirzi.
\newblock Self-instruct: Aligning language model with self generated
  instructions.
\newblock {\em arXiv preprint arXiv:2212.10560}, 2022.

\bibitem{alpaca}
Rohan Taori, Ishaan Gulrajani, Tianyi Zhang, Yann Dubois, Xuechen Li, Carlos
  Guestrin, Percy Liang, and Tatsunori~B. Hashimoto.
\newblock Stanford alpaca: An instruction-following llama model.
\newblock \url{https://github.com/tatsu-lab/stanford_alpaca}, 2023.

\bibitem{anandgpt4all}
Yuvanesh Anand, Zach Nussbaum, Brandon Duderstadt, Benjamin Schmidt, and Andriy
  Mulyar.
\newblock Gpt4all: Training an assistant-style chatbot with large scale data
  distillation from gpt-3.5-turbo.
\newblock \url{https://github.com/nomic-ai/gpt4all}, 2023.

\bibitem{peng2023instruction}
Baolin Peng, Chunyuan Li, Pengcheng He, Michel Galley, and Jianfeng Gao.
\newblock Instruction tuning with gpt-4.
\newblock {\em arXiv preprint arXiv:2304.03277}, 2023.

\bibitem{vicuna}
Vicuna.
\newblock Vicuna: An open-source chatbot impressing gpt-4 with 90\%* chatgpt
  quality.
\newblock \url{https://vicuna.lmsys.org/}, 2023.

\bibitem{clark1991grounding}
Herbert~H. Clark and Susan~E. Brennan.
\newblock Grounding in communication.
\newblock In Lauren Resnick, Levine B., M.~John, Stephanie Teasley, and D.,
  editors, {\em Perspectives on Socially Shared Cognition}, pages 13--1991.
  American Psychological Association, 1991.

\bibitem{falcon40b}
Ebtesam Almazrouei, Hamza Alobeidli, Abdulaziz Alshamsi, Alessandro Cappelli,
  Ruxandra Cojocaru, Merouane Debbah, Etienne Goffinet, Daniel Heslow, Julien
  Launay, Quentin Malartic, Badreddine Noune, Baptiste Pannier, and Guilherme
  Penedo.
\newblock {Falcon-40B}: an open large language model with state-of-the-art
  performance, 2023.

\bibitem{askell2021general}
Amanda Askell, Yuntao Bai, Anna Chen, Dawn Drain, Deep Ganguli, Tom Henighan,
  Andy Jones, Nicholas Joseph, Ben Mann, Nova DasSarma, Nelson Elhage, Zac
  Hatfield-Dodds, Danny Hernandez, Jackson Kernion, Kamal Ndousse, Catherine
  Olsson, Dario Amodei, Tom Brown, Jack Clark, Sam McCandlish, Chris Olah, and
  Jared Kaplan.
\newblock A general language assistant as a laboratory for alignment, 2021.

\bibitem{eval-harness}
Leo Gao, Jonathan Tow, Stella Biderman, Sid Black, Anthony DiPofi, Charles
  Foster, Laurence Golding, Jeffrey Hsu, Kyle McDonell, Niklas Muennighoff,
  Jason Phang, Laria Reynolds, Eric Tang, Anish Thite, Ben Wang, Kevin Wang,
  and Andy Zou.
\newblock A framework for few-shot language model evaluation, September 2021.

\bibitem{openai2023gpt4}
OpenAI.
\newblock Gpt-4 technical report, 2023.

\bibitem{chen2021codex}
Mark Chen, Jerry Tworek, Heewoo Jun, Qiming Yuan, Henrique~Ponde
  de~Oliveira~Pinto, Jared Kaplan, Harri Edwards, Yuri Burda, Nicholas Joseph,
  Greg Brockman, Alex Ray, Raul Puri, Gretchen Krueger, Michael Petrov, Heidy
  Khlaaf, Girish Sastry, Pamela Mishkin, Brooke Chan, Scott Gray, Nick Ryder,
  Mikhail Pavlov, Alethea Power, Lukasz Kaiser, Mohammad Bavarian, Clemens
  Winter, Philippe Tillet, Felipe~Petroski Such, Dave Cummings, Matthias
  Plappert, Fotios Chantzis, Elizabeth Barnes, Ariel Herbert-Voss,
  William~Hebgen Guss, Alex Nichol, Alex Paino, Nikolas Tezak, Jie Tang, Igor
  Babuschkin, Suchir Balaji, Shantanu Jain, William Saunders, Christopher
  Hesse, Andrew~N. Carr, Jan Leike, Josh Achiam, Vedant Misra, Evan Morikawa,
  Alec Radford, Matthew Knight, Miles Brundage, Mira Murati, Katie Mayer, Peter
  Welinder, Bob McGrew, Dario Amodei, Sam McCandlish, Ilya Sutskever, and
  Wojciech Zaremba.
\newblock Evaluating large language models trained on code, 2021.

\bibitem{Detoxify}
Laura Hanu and {Unitary team}.
\newblock Detoxify.
\newblock Github. https://github.com/unitaryai/detoxify, 2020.

\bibitem{glaese2022improving}
Amelia Glaese, Nat McAleese, Maja Trębacz, John Aslanides, Vlad Firoiu, Timo
  Ewalds, Maribeth Rauh, Laura Weidinger, Martin Chadwick, Phoebe Thacker, Lucy
  Campbell-Gillingham, Jonathan Uesato, Po-Sen Huang, Ramona Comanescu, Fan
  Yang, Abigail See, Sumanth Dathathri, Rory Greig, Charlie Chen, Doug Fritz,
  Jaume~Sanchez Elias, Richard Green, Soňa Mokrá, Nicholas Fernando, Boxi Wu,
  Rachel Foley, Susannah Young, Iason Gabriel, William Isaac, John Mellor,
  Demis Hassabis, Koray Kavukcuoglu, Lisa~Anne Hendricks, and Geoffrey Irving.
\newblock Improving alignment of dialogue agents via targeted human judgements,
  2022.

\bibitem{kryscinski2019evaluating}
Wojciech Kry{\'s}ci{\'n}ski, Bryan McCann, Caiming Xiong, and Richard Socher.
\newblock Evaluating the factual consistency of abstractive text summarization.
\newblock {\em arXiv preprint arXiv:1910.12840}, 2019.

\bibitem{shuster2021retrieval}
Kurt Shuster, Spencer Poff, Moya Chen, Douwe Kiela, and Jason Weston.
\newblock Retrieval augmentation reduces hallucination in conversation.
\newblock {\em arXiv preprint arXiv:2104.07567}, 2021.

\bibitem{henderson2018ethical}
Peter Henderson, Koustuv Sinha, Nicolas Angelard-Gontier, Nan~Rosemary Ke,
  Genevieve Fried, Ryan Lowe, and Joelle Pineau.
\newblock Ethical challenges in data-driven dialogue systems.
\newblock In {\em Proceedings of the 2018 AAAI/ACM Conference on AI, Ethics,
  and Society}, pages 123--129, 2018.

\bibitem{dixon2018measuring}
Lucas Dixon, John Li, Jeffrey Sorensen, Nithum Thain, and Lucy Vasserman.
\newblock Measuring and mitigating unintended bias in text classification.
\newblock In {\em Proceedings of the 2018 AAAI/ACM Conference on AI, Ethics,
  and Society}, pages 67--73, 2018.

\bibitem{perez2022discovering}
Ethan Perez, Sam Ringer, Kamilė Lukošiūtė, Karina Nguyen, Edwin Chen, Scott
  Heiner, Craig Pettit, Catherine Olsson, Sandipan Kundu, Saurav Kadavath, Andy
  Jones, Anna Chen, Ben Mann, Brian Israel, Bryan Seethor, Cameron McKinnon,
  Christopher Olah, Da~Yan, Daniela Amodei, Dario Amodei, Dawn Drain, Dustin
  Li, Eli Tran-Johnson, Guro Khundadze, Jackson Kernion, James Landis, Jamie
  Kerr, Jared Mueller, Jeeyoon Hyun, Joshua Landau, Kamal Ndousse, Landon
  Goldberg, Liane Lovitt, Martin Lucas, Michael Sellitto, Miranda Zhang, Neerav
  Kingsland, Nelson Elhage, Nicholas Joseph, Noemí Mercado, Nova DasSarma,
  Oliver Rausch, Robin Larson, Sam McCandlish, Scott Johnston, Shauna Kravec,
  Sheer~El Showk, Tamera Lanham, Timothy Telleen-Lawton, Tom Brown, Tom
  Henighan, Tristan Hume, Yuntao Bai, Zac Hatfield-Dodds, Jack Clark, Samuel~R.
  Bowman, Amanda Askell, Roger Grosse, Danny Hernandez, Deep Ganguli, Evan
  Hubinger, Nicholas Schiefer, and Jared Kaplan.
\newblock Discovering language model behaviors with model-written evaluations,
  2022.

\bibitem{weidinger2021ethical}
Laura Weidinger, John Mellor, Maribeth Rauh, Conor Griffin, Jonathan Uesato,
  Po-Sen Huang, Myra Cheng, Mia Glaese, Borja Balle, Atoosa Kasirzadeh, et~al.
\newblock Ethical and social risks of harm from language models.
\newblock {\em arXiv preprint arXiv:2112.04359}, 2021.

\bibitem{christiano2023deep}
Paul Christiano, Jan Leike, Tom~B. Brown, Miljan Martic, Shane Legg, and Dario
  Amodei.
\newblock Deep reinforcement learning from human preferences, 2023.

\bibitem{stiennon2022learning}
Nisan Stiennon, Long Ouyang, Jeff Wu, Daniel~M. Ziegler, Ryan Lowe, Chelsea
  Voss, Alec Radford, Dario Amodei, and Paul Christiano.
\newblock Learning to summarize from human feedback, 2022.

\bibitem{tidemanIndependenceClonesCriterion1987}
T.~N. Tideman.
\newblock Independence of clones as a criterion for voting rules.
\newblock {\em Social Choice and Welfare}, 4(3):185--206, September 1987.

\bibitem{chatgptbias2023}
David Rozado.
\newblock The political biases of chatgpt.
\newblock {\em Social Sciences}, 12(3):148, Mar 2023.

\end{thebibliography}

\newpage

\appendix
\addcontentsline{toc}{section}{Appendix}
\part{Appendix} % Start the appendix part
\parttoc % Insert the appendix TOC

\section{Contributor Guidelines}
\label{app:contributor_guidelines}

We provide the guidelines presented to the users for the creation of the dataset.

\begin{verbatim}
# Guidelines

Below is a list of guidelines that should be adhered to for each possible task
available when building the dataset. To see some examples of how the guidelines
can be applied, visit the examples document.

Please consider checking out our survey
[here](https://forms.gle/vBW7b2kMzjCoehkH9). You can use it to rate each
guideline and leave feedback for each task.

If you have further suggestions to improve any of our guidelines, or want to add
more examples, create a pull request or suggest them on our
[GitHub](https://github.com/LAION-AI/Open-Assistant).

## 1. General rules

- Always make sure to read and understand the guidelines to each task before
  fulfilling it.
- Try to follow the guidelines as closely as possible.
- If you are unsure whether a message violates a guidelines, contact us at our
  Discord.
- Use the thumbs-up/thumbs-down system to further mark messages that are of high
  or low quality.

## 2. Providing an assistant reply {#assistant-reply}

### Do:

- Remain polite and treat the user with respect, even when not given the same
  courtesy.
- Talk in a friendly and approachable manner, unless specifically requested
  otherwise.
- Present only information that has been verified by credible sources that can
  be backed up, unless specifically requested otherwise.
- Make sure the user is aware when given unverified information.
- Inform the user about the potential dangers when being asked for advice
  regarding a topic with high risk, such as medicine, law or chemistry.
- When being asked about a high-risk topic, make sure the user knows that as a
  language model, the assistant is susceptible to producing incorrect
  information, and that no actions should be taken regarding the assistant reply
  without the opinion of a professional.
- When being asked to give an opinion as the default persona of the assistant,
  make sure to bring up at least 2 common viewpoints and ensure that these
  aren't expressed as the opinions of the assistant.
  - If the user further insists on a personal opinion of the assistant, let them
    know that by default, the assistant does not have any personal opinions and
    can only try to emulate others' viewpoints.
- Ask for clarification if it's unclear what the user is asking for.
- Use paragraphs and line breaks to make larger replies more readable.
- Make use of [Markdown syntax](https://www.markdownguide.org/basic-syntax) to
  better format lists, tables or blocks of code.
  - If you are using a codeblock to write code in a particular language, specify
    it to enable
    [syntax highlighting]
    (https://www.markdownguide.org/extended-syntax/#syntax-highlighting).
    You can find all supported abbreviations
    [here](https://github.com/jincheng9/markdown_supported_languages
    #heres-a-full-list-of-supported-languages).
- Be consistent in the style and tone of the assistant.

### Don't:

- Copy and paste text from other sources without editing. **This includes
  ChatGPT.** 
- Supply text that violates the law of Germany, UK, USA, or your country of
  residence.
- Write content encouraging:
  - Violence
  - Violation of the rights of a third party
  - Pedophilia
- Provide the user with information that could be used for self-harm if there is
  plausible suspicion of intent to self-harm.
- Provide personal information of third parties that isn't publicly available.
- Ask for personal information unless it is relevant to the issue and can't be
  used to determine the identity of the user, such as country of residence or
  occupation. The user should be allowed to refuse to give up any information.
- Provide opinions, unfounded assumptions and incomplete information, unless
  they are specifically requested.
- Purposefully curate information to guide the conclusion, i.e. don't hide facts
  to present a particular narrative.
- Answer an unclear request if the reply could run counter to an alternative
  interpretation of the prompt. Ask the user to elaborate or rephrase instead.
- Dodge a question, unless it violates a guideline.
- Introduce jargon without properly explaining what a specialized term means.
  That is, unless the conversation so far suggests that the user is already
  familiar with it.
- Leave typos or grammatical errors in the assistant replies, unless
  specifically requested to do so.
- Overload the user with too much information. Keep replies concise, but include
  further details that relate to and expand upon the user's request.
- Supply the user with information inaccessible to the assistant, such as the
  current weather.
- Reply in a language different from the one intended for the dataset, unless
  specifically requested to do so.

## 3. Providing an initial prompt or user reply {#user-reply}

### Do:

- Ask questions that reflect real-life situations and needs.
- Ask questions that might be directed towards search engines or specialists.
- Make requests that encourage lateral thinking and/or require specialized
  knowledge.
- Use a mix between questions that are straightforward and questions without a
  clear answer.
- Introduce a variety in prompts by using different phrasing, degrees of
  politeness or amount of context given.
- Consider the previous replies and prompts that lead up to the current one.
- Try to build upon the topic and ask a sensible follow-up question when
  replying to the assistant.

### Don't:

- Write prompts without a clear request.
- Supply text that violates the law of Germany, UK, USA, or your country of
  residence.
- Make requests that override the original purpose of the assistant, i.e.
  jailbreak the model.
- Make requests that leave the assistant with no other choice but to refuse in
  order to avoid the generation of harmful content.
- Submit a prompt similar or identical to a prompt you previously submitted.
- Change the topic of a conversation without prefacing it accordingly when
  replying to the assistant.
- Leave typos and grammatical errors in the prompt.
- Reply in a language different from the one intended for the dataset, unless
  the context of the conversation requires it.

## 4. Classifying an assistant reply {#classifying-assistant}

### Do:

- Rate every criteria of each reply, unless it can't be discerned because it is
  spam or inappropriate.
- Judge quality based on how well the reply adheres to the guidelines. Factual
  accuracy and helpfulness are first and foremost.
- Make sure to read the reply thoroughly.
- Use the [label explanations](#label-explanation) to determine which labels
  apply to the reply.
- Research to make sure whether the reply is factually accurate.
- Skip a classification if you are unable to determine the validity of reply.

### Don't:

- Judge quality based on personal beliefs. Assuming an opinion was warranted,
  fulfills the users request and doesn't violate any guidelines, it should not
  impact the rating of the reply.
- Skip a label just because the reply is spam. Each label can help the model
  improve.
- Rate a reply if you are unsure if it factually accurate or satisfies the
  request of the user.

## 5. Classifying an initial prompt or user reply {#classifying-user}

### Do:

- Rate every criteria of each prompt, unless it can't be discerned because it is
  spam or inappropriate.
- Judge quality based on how well the prompt adheres to the guidelines.
- Make sure to read the prompt thoroughly.
- Use the [label explanations](#label-explanation) to determine which labels
  apply to the prompt.

### Don't:

- Judge quality based on personal beliefs. The opinion of the user should not
  impact the rating of the prompt.
- Skip a label just because the reply is spam. Each label can help the model
  improve.

## 6. Ranking assistant replies {#ranking-assistant}

### Do:

- Make sure to read every available reply.
- Think about which reply best satisfies the request of the user.
- Rank replies based on how well they adhere to the guidelines. Factual accuracy
  and helpfulness are first and foremost.
- Penalize replies that fail to provide adequate warnings or caveats.
- Penalize replies that are difficult to read due to a lack of formatting,
  capitalization or other errors.
- Penalize replies if the requested information is obfuscated by superfluous
  details that make up a large part of the message.
- Rank replies that admit to not knowing the answer below factually correct, but
  above factually incorrect replies.

### Don't:

- Rank replies based on personal beliefs. Assuming an opinion was warranted,
  fulfills the users request and doesn't violate any guidelines, it should not
  impact the rating of the reply.
- Rank replies based on how long and short they are - instead, find out which
  reply best answers the query of the user.

## 7. Label Explanation {#label-explanation}

**Is the message spam?**

Is the message an unrelated piece of text that doesn't contribute to the
previous prompt in any way? If it's an initial prompt, is it something unlikely
to be used to initiate a conversation with the assistant?

**Is it a bad reply, as an answer to the prompt task?**

Does the assistant reply fail to engage the request of the user or does it
contain incorrect information? Does it ignore relevant parts of the prompt?

### Tags

**Not Target Language**: Unless specifically requested, is the language used
different from the target language of the dataset?

**Not Appropriate**: In case of a user prompt, is it missing a clear request?

**Contains PII**: Does the message contain personally identifying information
(PII) that isn't publicly available and can be used to determine the identity of
the user or a private third party?

**Hate Speech**: Does the message contain prejudice towards a group classified
as protected according to the law of Germany, UK, USA or your country of
residence?

**Sexual Content**: Does the message contain or request sexually explicit
content, such as descriptions of sexual acts?

### Scales

**Low Quality/High Quality**: How well does the message adhere to the
guidelines? For assistant replies, factual accuracy and helpfulness are first
and foremost.

**Serious/Humorous**: Does the message contain sarcasm, a play on words or other
humorous embellishments?

**Ordinary/Creative**: How unlikely is the question to be asked? Is it
challenging the assistant with a problem that requires creative thinking? Does
the question include lots of context? On the assistant's side, is the reply
straightforward, or does it contain a unique approach to a problem?

**Rude/Polite**: How polite is the message towards the recipient? Does it use
words like "please" or is it dismissive, rude or insulting towards the other
side of the conversation?

**Violent/Harmless**: Does the message contain descriptions of explicit violent
acts? Is it glorifying, encouraging or downplaying violent behavior?

\end{verbatim}

\section{Ranking merging}
\label{app:ranking}

Reinforcement learning from human feedback~\cite{christiano2023deep,stiennon2022learning} comprises a set of techniques that all aim to optimize the output distribution of a language model using the preference structure provided by human rankers.
To get a preference structure that is well aligned to users, we cannot just rely on the opinions of individual rankers, due to the high variance in human preferences.
Since our objective is to collect data for a generally capable digital assistant, every ranking of possible responses is performed by K independent rankers, (see, section~\ref{sec:single-state-collection}).
Once this is done, we need to fuse these K individual opinions into one consensus opinion usable in training preference models.\par
We perform this preference fusion by treating it as a ranked-voting problem, whose objective it is to maintain the preferences as faithfully as possible.
The method chosen for this is known as "ranked pairs" or "Tideman's method"~\cite{tidemanIndependenceClonesCriterion1987}.
Simplified, this method creates a sorted list of "winners" according to the strength of the preference of one element over the others.
The way the preference strength is measured is by considering all preference pairs in the input votes:
For example, if the votes are two times $A>B>C$ and one time $B>A>C$, this would mean that the pair $A>B$ exists 2 times, while $A>C$ exists three times.
The method then orders the winners by winning strength, i.e. here $A>C$ would happen before $A>B$, and constructs a directed graph using the preferences, i.e. $A>C$ would become an edge $A\to C$.
Edges are added one by one according to their weight (higher first), as long as no cycle is produced.
If the edge would produce a cycle, it is skipped as the corresponding dominant preferences are already in the graph.\par
The resulting directed acyclic graph can be turned into a preference structure by recursively removing source nodes and adding them to the back of a preference list, since source nodes are, by construction, not preferred over any other item left in the graph.
In practice, one can speed up the construction by not explicitly constructing the preference graph and fusing the graph con- and destruction.

\section{Web application for data collection}
\label{app:webapp}

\begin{figure}[!htb]
    \centering
    \includegraphics[width=0.55\linewidth]{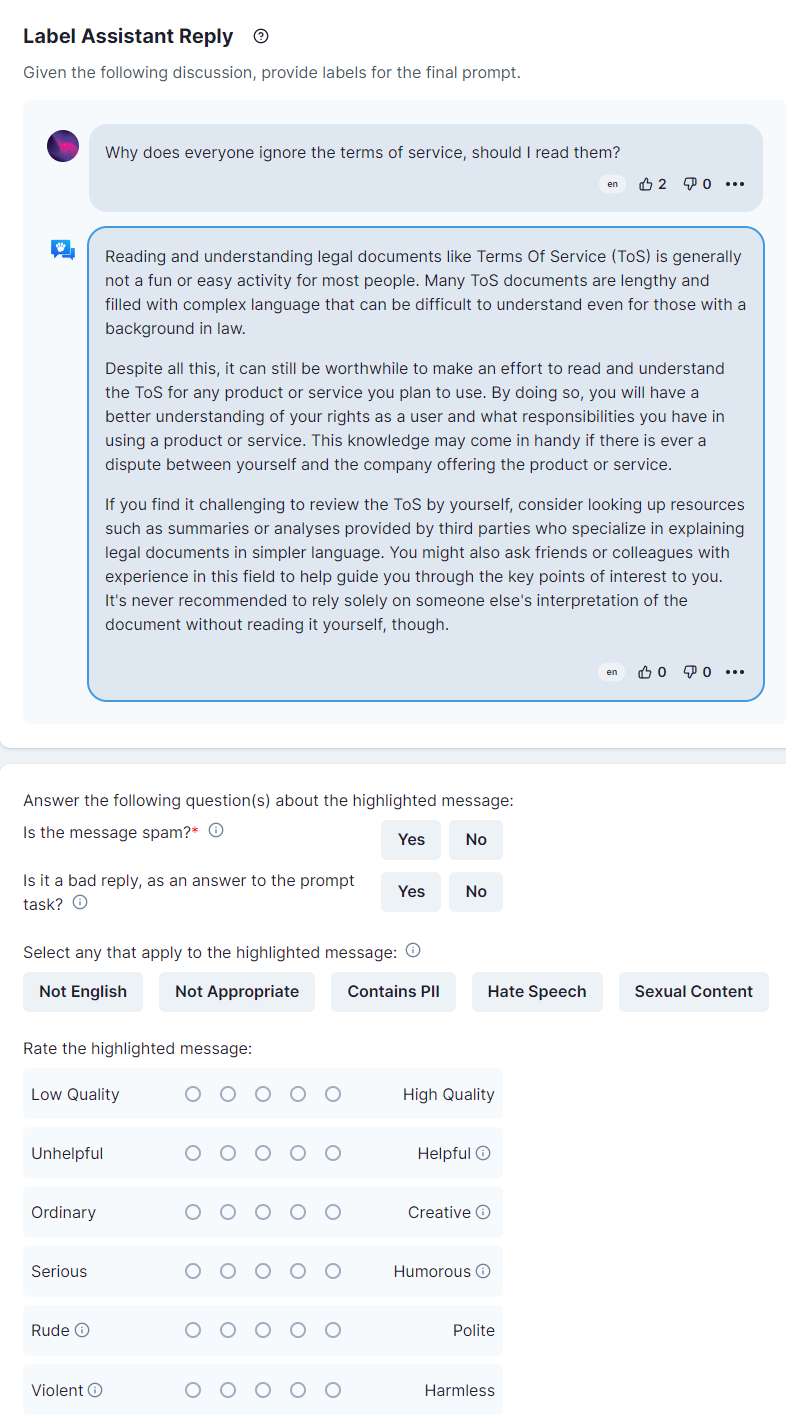}
    \caption{A preview of the page for labelling tasks. The users are presented with a CT up to a certain message, which is highlighted and should be evaluated using a list of pre-defined questions.}
    \label{fig:web_labelling}
\end{figure}

\begin{figure}[!htb]
    \centering
    \includegraphics[width=0.5\textwidth]{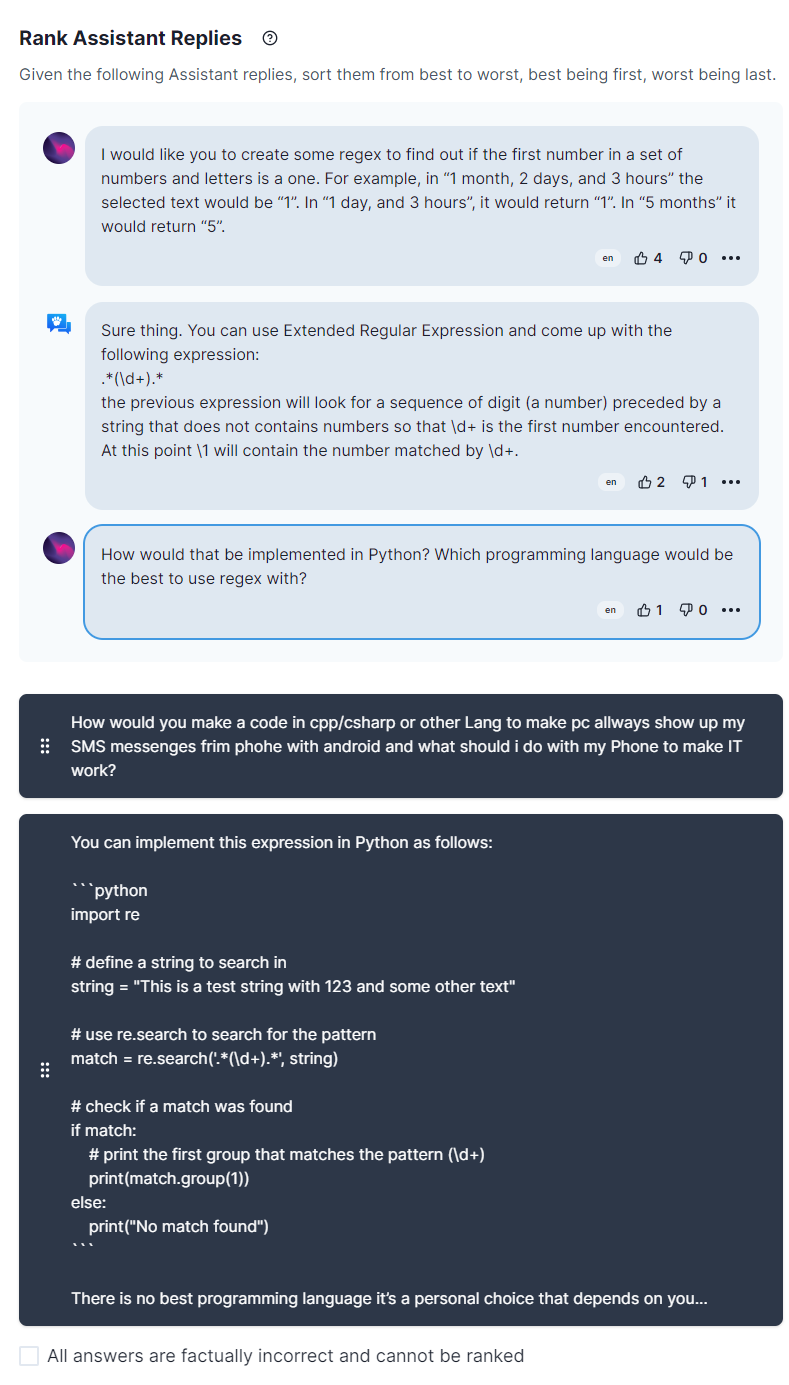}
    \caption{A preview of the page for ranking assistant replies. Users are provided with CT up to a message, and a couple of responses that should be ranked according to how good they answer the given message. The interaction is drag \& drop based. Additionally, users can choose to mark all answers as factually incorrect.}
    \label{fig:web_rank_assistant_replies}
\end{figure}

\begin{figure}[!htb]
    \centering
    \includegraphics[width=0.6\textwidth]{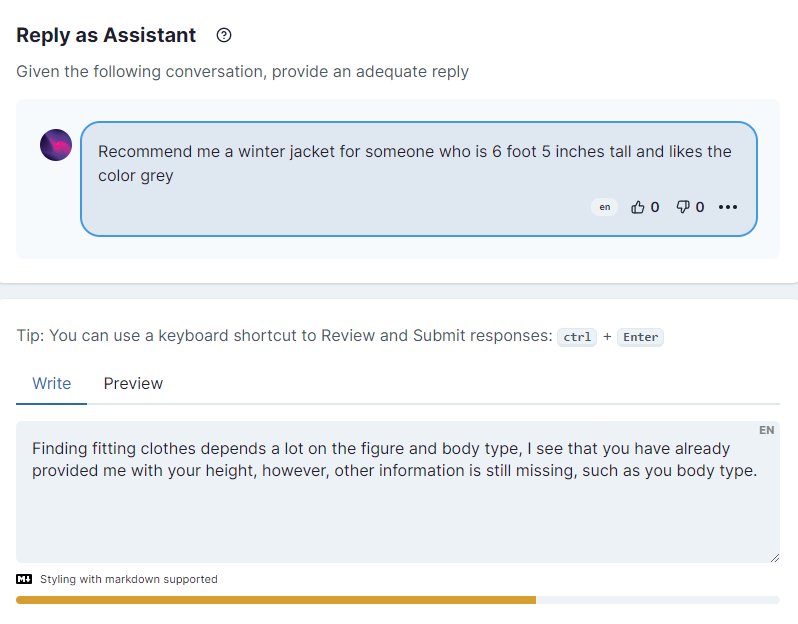}
    \caption{A preview of the page for replying as an assistant. Users are provided with a CT up to a prompter message, and they should provide a response to it. In this example, the CT contains only one message. Users can use Markdown in their responses. An additional progress bar is added below the input field to incentivize longer messages.}
    \label{fig:web_reply_assistant}
\end{figure}

Figures \ref{fig:web_labelling}, \ref{fig:web_rank_assistant_replies}, and \ref{fig:web_reply_assistant} show the data collection platform UI interface for labelling, reply input, and ranking, respectively. The UI also allows the user to choose a random task type, the task is automatically chosen according the previously discussed heuristics, which allows for more variation and engagement.

The UI did grow and change over time as the project developed, so the experience of early users was less refined as shown here.

\FloatBarrier
\section{Filtering ChatGPT Inputs}
The widespread use of ChatGPT at the time of our data collection meant that some of the inputs could be performed using ChatGPT, rather than being fully created by humans.
Our guidelines specifically target this, explicitly discouraging contributors from copy-pasting responses generated by other AI models.
Users who were found to post ChatGPT-generated responses were banned, and their contributions were deleted. Furthermore, we used multiple automatic tests to catch such cases.
For instance, we searched for and removed messages that contained text such as "as a large language model" or "knowledge cutoff after September 2021".
Moreover, users were encouraged to up-vote and down-vote responses they came across from other users, which also helped weeding out low-quality, generic, (possibly AI generated) responses.
We note that these mechanisms cannot remove all ChatGPT-generated content, and more post-hoc filtering may be necessary.

\FloatBarrier
\section{Online Survey Results}

\begin{figure}[!htb]
    \centering
    \includegraphics[width=\linewidth]{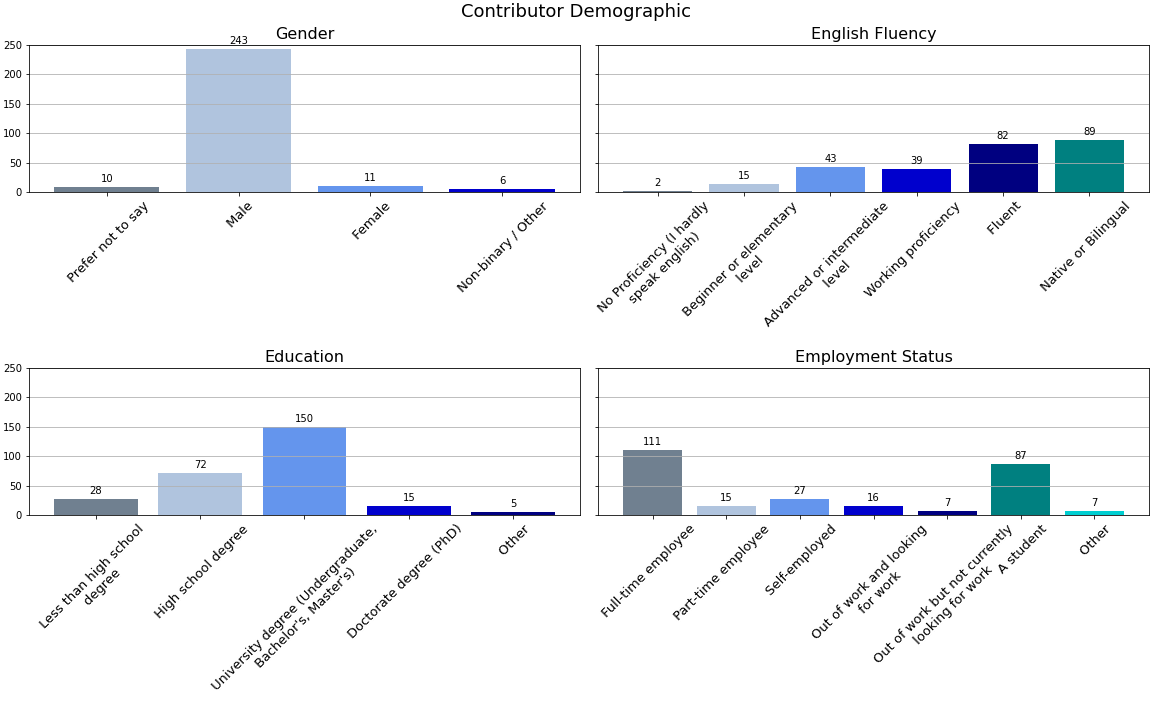}
    \caption{Demography of 270 respondents}
    \label{fig:demography}
\end{figure}

\begin{figure}
    \centering
    \includegraphics[width=\linewidth]{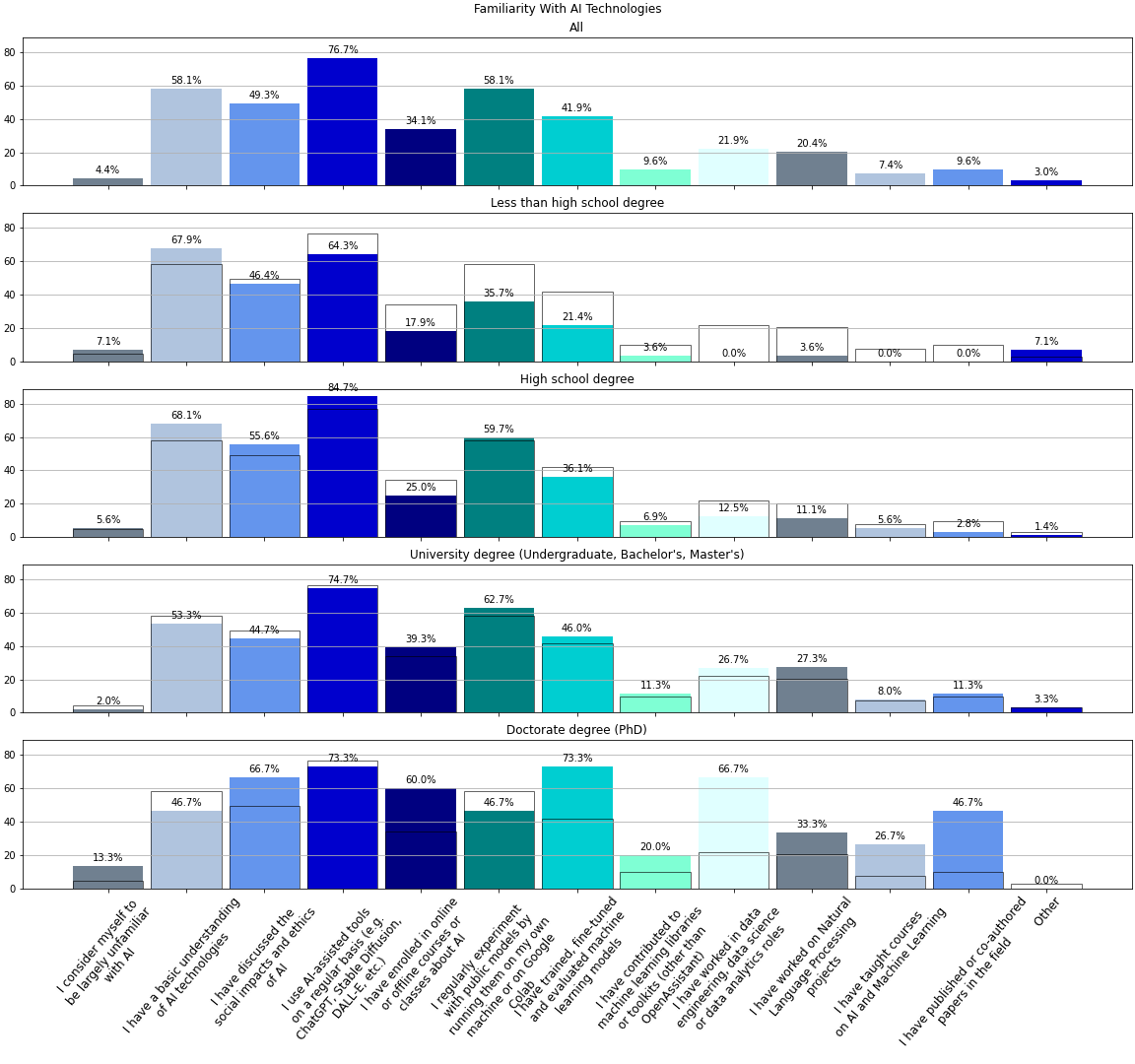}
    \caption{Familiarity with AI based on level of education. The average values are shown as outlines in the bar chart.}
    \label{fig:familiarityAI}
\end{figure}

\begin{figure}
    \centering
    \includegraphics[width=\linewidth]{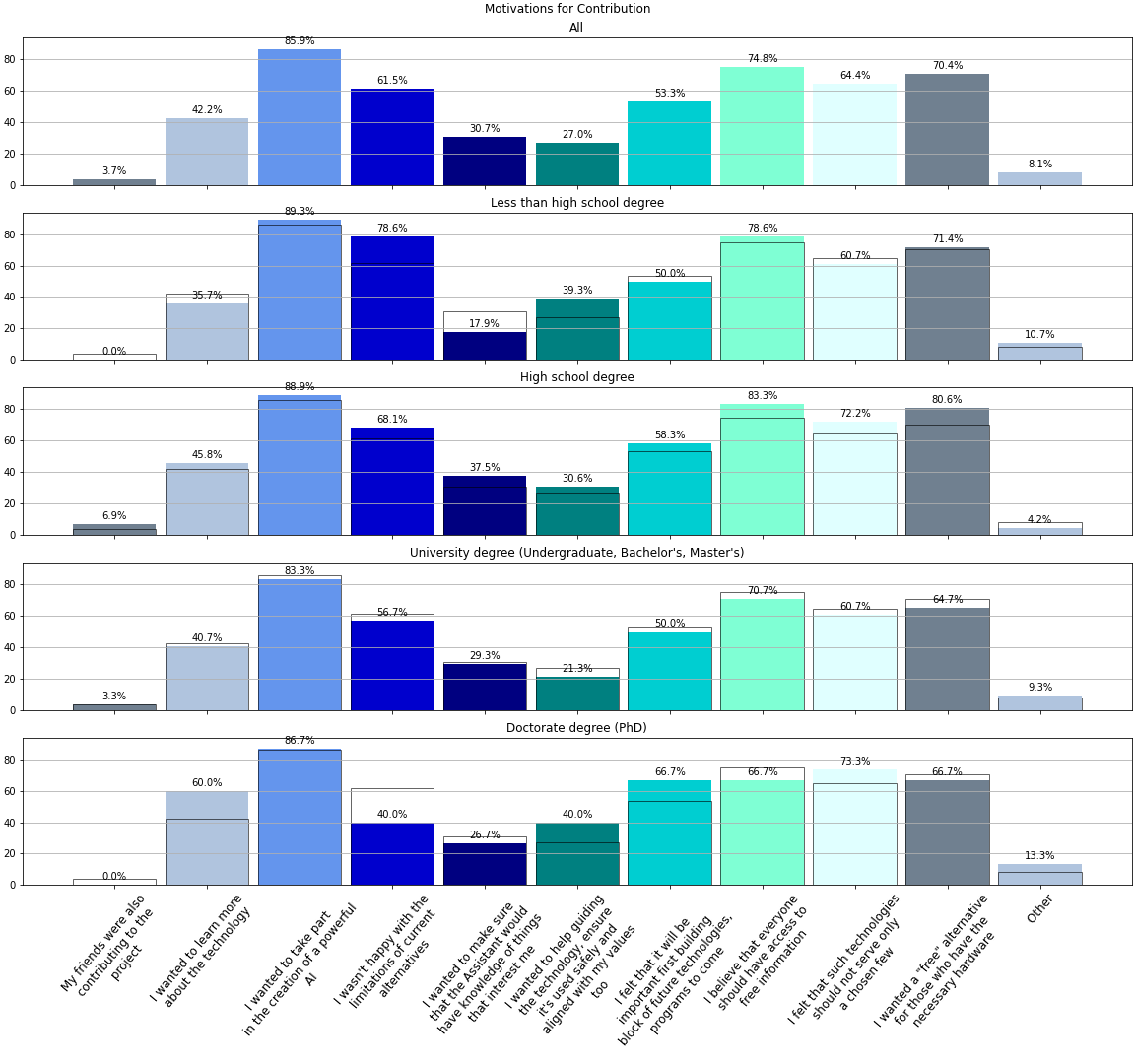}
    \caption{Motivations for contribution based on level of education. The average values are shown as outlines in the bar chart.}
    \label{fig:Motivation}
\end{figure}

\begin{figure}
    \centering
    \includegraphics[width=\linewidth]{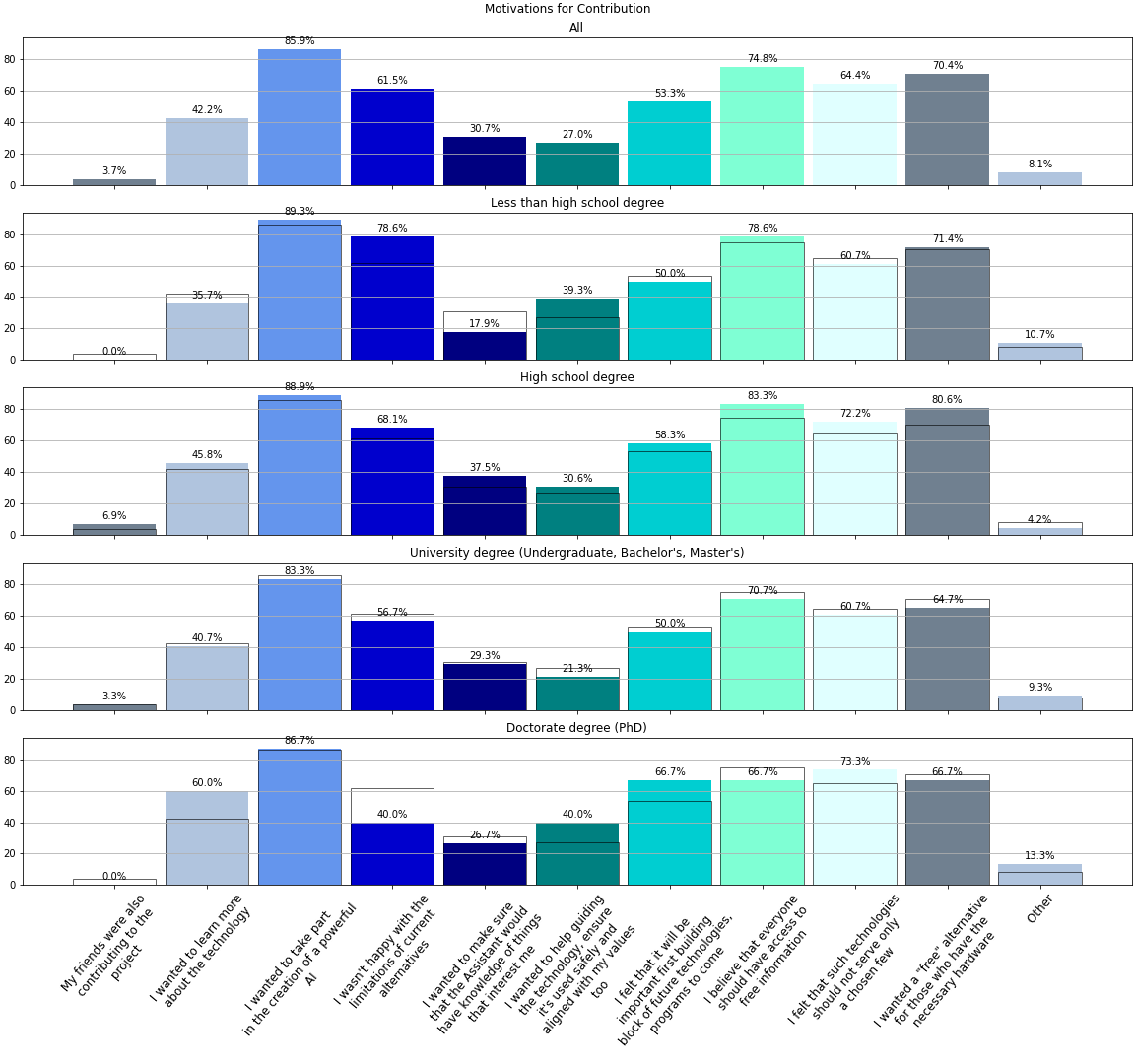}
    \caption{Planned personal use-cases for OpenAssistant. The average values are shown as outlines in the bar chart.}
    \label{fig:UseCases}
\end{figure}

We asked users to provide feedback based on the overall experience, while taking part in the data collection process. Results are provided in Tables~\ref{tab:LikertSatisfaction},~\ref{tab:contributePrevious} and Fig.~\ref{fig:demography},~\ref{fig:familiarityAI},~\ref{fig:Motivation},~\ref{fig:UseCases}.

\begin{table}[!htb]
    \centering

    \begin{tabular}{lrrrr}
    
    \multicolumn{2}{c}{It was clear from the website what to do and how to contribute messages to the training data.} \\
    \hline
    \multicolumn{1}{l}{Strongly agree} & \multicolumn{1}{r}{33.70\%}  \\
    \multicolumn{1}{l}{Agree} & \multicolumn{1}{r}{47.04\%}  \\
    \multicolumn{1}{l}{Neither agree nor disagree} & \multicolumn{1}{r}{11.48\%}  \\
    \multicolumn{1}{l}{Disagree} & \multicolumn{1}{r}{6.67\%}  \\
    \multicolumn{1}{l}{Strongly disagree} & \multicolumn{1}{r}{1.11\%}  \\
    
    \multicolumn{2}{c}{I've felt I could always ask for help from the community and the moderators.} \\
    \hline
    \multicolumn{1}{l}{Strongly agree} & \multicolumn{1}{r}{30.74\%}  \\
    \multicolumn{1}{l}{Agree} & \multicolumn{1}{r}{27.78\%}  \\
    \multicolumn{1}{l}{Neither agree nor disagree} & \multicolumn{1}{r}{31.85\%}  \\
    \multicolumn{1}{l}{Disagree} & \multicolumn{1}{r}{7.41\%}  \\
    \multicolumn{1}{l}{Strongly disagree} & \multicolumn{1}{r}{2.22\%}  \\
    
    \multicolumn{2}{c}{I found the tasks enjoyable and engaging.} \\
    \hline
    \multicolumn{1}{l}{Strongly agree} & \multicolumn{1}{r}{20.00\%}  \\
    \multicolumn{1}{l}{Agree} & \multicolumn{1}{r}{42.22\%}  \\
    \multicolumn{1}{l}{Neither agree nor disagree} & \multicolumn{1}{r}{27.78\%}  \\
    \multicolumn{1}{l}{Disagree} & \multicolumn{1}{r}{8.15\%}  \\
    \multicolumn{1}{l}{Strongly disagree} & \multicolumn{1}{r}{1.85\%}  \\
    
    \multicolumn{2}{c}{I found the tasks repetitive.} \\
    \hline
    \multicolumn{1}{l}{Strongly agree} & \multicolumn{1}{r}{11.48\%}  \\
    \multicolumn{1}{l}{Agree} & \multicolumn{1}{r}{30.00\%}  \\
    \multicolumn{1}{l}{Neither agree nor disagree} & \multicolumn{1}{r}{37.04\%}  \\
    \multicolumn{1}{l}{Disagree} & \multicolumn{1}{r}{18.15\%}  \\
    \multicolumn{1}{l}{Strongly disagree} & \multicolumn{1}{r}{3.33\%}  \\
    
    \multicolumn{2}{c}{While doing rating or ranking tasks, I found the messages to be really high quality.} \\
    \hline
    \multicolumn{1}{l}{Strongly agree} & \multicolumn{1}{r}{16.67\%}  \\
    \multicolumn{1}{l}{Agree} & \multicolumn{1}{r}{42.59\%}  \\
    \multicolumn{1}{l}{Neither agree nor disagree} & \multicolumn{1}{r}{30.00\%}  \\
    \multicolumn{1}{l}{Disagree} & \multicolumn{1}{r}{9.26\%}  \\
    \multicolumn{1}{l}{Strongly disagree} & \multicolumn{1}{r}{1.48\%}  \\
    
    \multicolumn{2}{c}{Overall, I’m glad I have contributed to OpenAssistant.} \\
    \hline
    \multicolumn{1}{l}{Strongly agree} & \multicolumn{1}{r}{81.11\%}  \\
    \multicolumn{1}{l}{Agree} & \multicolumn{1}{r}{14.44\%}  \\
    \multicolumn{1}{l}{Neither agree nor disagree} & \multicolumn{1}{r}{2.96\%}  \\
    \multicolumn{1}{l}{Disagree} & \multicolumn{1}{r}{0.74\%}  \\
    \multicolumn{1}{l}{Strongly disagree} & \multicolumn{1}{r}{0.74\%}  \\
    
    \end{tabular}
    \caption{User Satisfaction Survey}
    \label{tab:LikertSatisfaction}
\end{table}

\begin{table}[!htb]
    \centering
    \begin{tabular}{lr}
    \toprule
     &  \\
    Have you contributed to other community projects besides this one? &  \\
    \midrule
    No, this is my first time contributing & 111 \\
    Yes I have contributed to a few projects & 110 \\
    Yes, I have contributed to multiple open source projects & 44 \\
    Prefer not to say & 5 \\
    \bottomrule
    \end{tabular}
    \caption{Previous Contributions}
    \label{tab:contributePrevious}
\end{table}

\FloatBarrier
\section{Word Cloud}
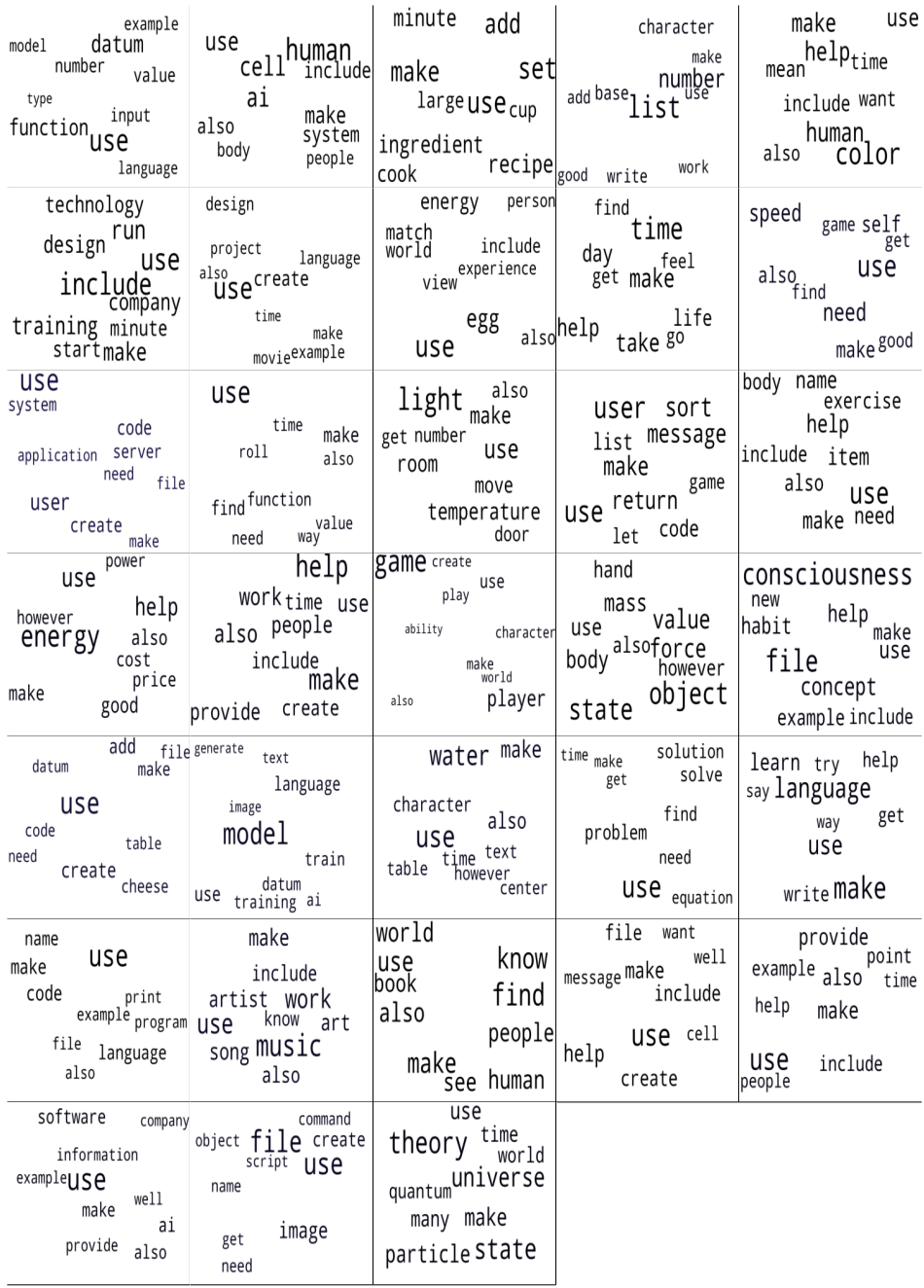
\begin{figure}[!htb]
    \centering
\begin{tikzpicture}
\node[anchor=north west,inner sep=0pt] at (0.0,0.0) {\includegraphics[width=2.5cm,height=2.5cm]{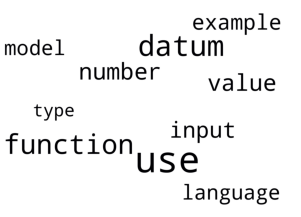}};
\draw[draw=black,line width=0.1pt] (2.5,0.0) -- (2.5,-2.5);
\draw[draw=black,line width=0.1pt] (0.0,-2.5) -- (2.5,-2.5);
\node[anchor=north west,inner sep=0pt] at (2.5,0.0) {\includegraphics[width=2.5cm,height=2.5cm]{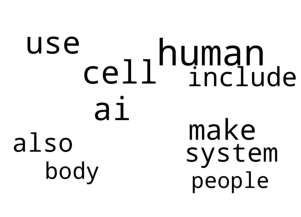}};
\draw[draw=black,line width=0.1pt] (5.0,0.0) -- (5.0,-2.5);
\draw[draw=black,line width=0.1pt] (2.5,-2.5) -- (5.0,-2.5);
\node[anchor=north west,inner sep=0pt] at (5.0,0.0) {\includegraphics[width=2.5cm,height=2.5cm]{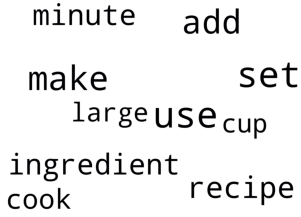}};
\draw[draw=black,line width=0.1pt] (7.5,0.0) -- (7.5,-2.5);
\draw[draw=black,line width=0.1pt] (5.0,-2.5) -- (7.5,-2.5);
\node[anchor=north west,inner sep=0pt] at (7.5,0.0) {\includegraphics[width=2.5cm,height=2.5cm]{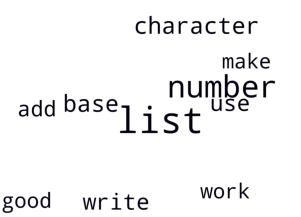}};
\draw[draw=black,line width=0.1pt] (10.0,0.0) -- (10.0,-2.5);
\draw[draw=black,line width=0.1pt] (7.5,-2.5) -- (10.0,-2.5);
\node[anchor=north west,inner sep=0pt] at (10.0,0.0) {\includegraphics[width=2.5cm,height=2.5cm]{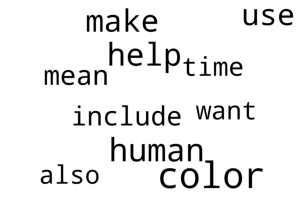}};
\draw[draw=black,line width=0.1pt] (10.0,-2.5) -- (12.5,-2.5);
\node[anchor=north west,inner sep=0pt] at (0.0,-2.5) {\includegraphics[width=2.5cm,height=2.5cm]{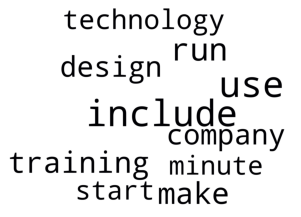}};
\draw[draw=black,line width=0.1pt] (2.5,-2.5) -- (2.5,-5.0);
\draw[draw=black,line width=0.1pt] (0.0,-5.0) -- (2.5,-5.0);
\node[anchor=north west,inner sep=0pt] at (2.5,-2.5) {\includegraphics[width=2.5cm,height=2.5cm]{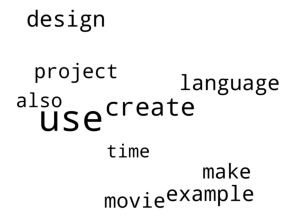}};
\draw[draw=black,line width=0.1pt] (5.0,-2.5) -- (5.0,-5.0);
\draw[draw=black,line width=0.1pt] (2.5,-5.0) -- (5.0,-5.0);
\node[anchor=north west,inner sep=0pt] at (5.0,-2.5) {\includegraphics[width=2.5cm,height=2.5cm]{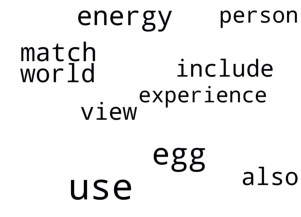}};
\draw[draw=black,line width=0.1pt] (7.5,-2.5) -- (7.5,-5.0);
\draw[draw=black,line width=0.1pt] (5.0,-5.0) -- (7.5,-5.0);
\node[anchor=north west,inner sep=0pt] at (7.5,-2.5) {\includegraphics[width=2.5cm,height=2.5cm]{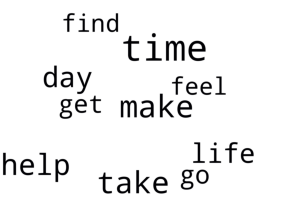}};
\draw[draw=black,line width=0.1pt] (10.0,-2.5) -- (10.0,-5.0);
\draw[draw=black,line width=0.1pt] (7.5,-5.0) -- (10.0,-5.0);
\node[anchor=north west,inner sep=0pt] at (10.0,-2.5) {\includegraphics[width=2.5cm,height=2.5cm]{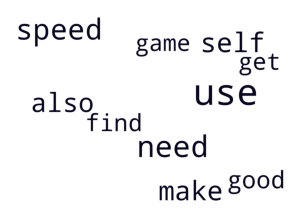}};
\draw[draw=black,line width=0.1pt] (10.0,-5.0) -- (12.5,-5.0);
\node[anchor=north west,inner sep=0pt] at (0.0,-5.0) {\includegraphics[width=2.5cm,height=2.5cm]{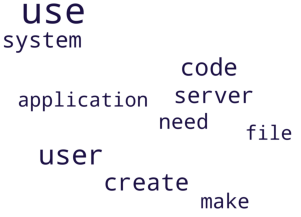}};
\draw[draw=black,line width=0.1pt] (2.5,-5.0) -- (2.5,-7.5);
\draw[draw=black,line width=0.1pt] (0.0,-7.5) -- (2.5,-7.5);
\node[anchor=north west,inner sep=0pt] at (2.5,-5.0) {\includegraphics[width=2.5cm,height=2.5cm]{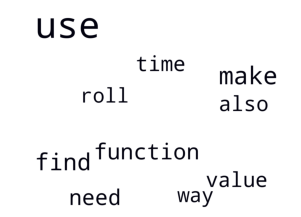}};
\draw[draw=black,line width=0.1pt] (5.0,-5.0) -- (5.0,-7.5);
\draw[draw=black,line width=0.1pt] (2.5,-7.5) -- (5.0,-7.5);
\node[anchor=north west,inner sep=0pt] at (5.0,-5.0) {\includegraphics[width=2.5cm,height=2.5cm]{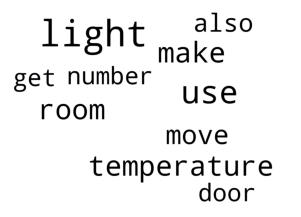}};
\draw[draw=black,line width=0.1pt] (7.5,-5.0) -- (7.5,-7.5);
\draw[draw=black,line width=0.1pt] (5.0,-7.5) -- (7.5,-7.5);
\node[anchor=north west,inner sep=0pt] at (7.5,-5.0) {\includegraphics[width=2.5cm,height=2.5cm]{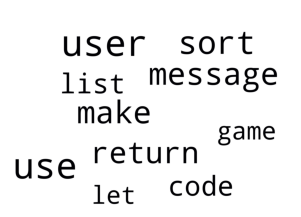}};
\draw[draw=black,line width=0.1pt] (10.0,-5.0) -- (10.0,-7.5);
\draw[draw=black,line width=0.1pt] (7.5,-7.5) -- (10.0,-7.5);
\node[anchor=north west,inner sep=0pt] at (10.0,-5.0) {\includegraphics[width=2.5cm,height=2.5cm]{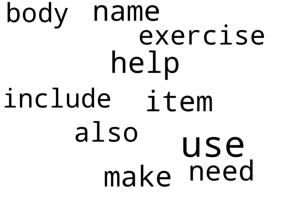}};
\draw[draw=black,line width=0.1pt] (10.0,-7.5) -- (12.5,-7.5);
\node[anchor=north west,inner sep=0pt] at (0.0,-7.5) {\includegraphics[width=2.5cm,height=2.5cm]{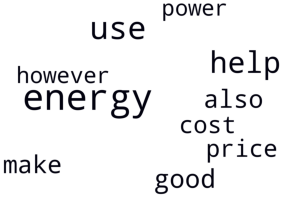}};
\draw[draw=black,line width=0.1pt] (2.5,-7.5) -- (2.5,-10.0);
\draw[draw=black,line width=0.1pt] (0.0,-10.0) -- (2.5,-10.0);
\node[anchor=north west,inner sep=0pt] at (2.5,-7.5) {\includegraphics[width=2.5cm,height=2.5cm]{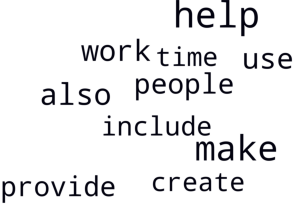}};
\draw[draw=black,line width=0.1pt] (5.0,-7.5) -- (5.0,-10.0);
\draw[draw=black,line width=0.1pt] (2.5,-10.0) -- (5.0,-10.0);
\node[anchor=north west,inner sep=0pt] at (5.0,-7.5) {\includegraphics[width=2.5cm,height=2.5cm]{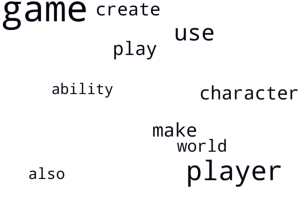}};
\draw[draw=black,line width=0.1pt] (7.5,-7.5) -- (7.5,-10.0);
\draw[draw=black,line width=0.1pt] (5.0,-10.0) -- (7.5,-10.0);
\node[anchor=north west,inner sep=0pt] at (7.5,-7.5) {\includegraphics[width=2.5cm,height=2.5cm]{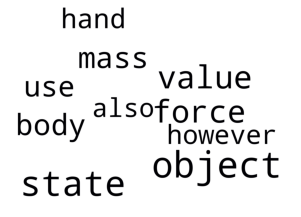}};
\draw[draw=black,line width=0.1pt] (10.0,-7.5) -- (10.0,-10.0);
\draw[draw=black,line width=0.1pt] (7.5,-10.0) -- (10.0,-10.0);
\node[anchor=north west,inner sep=0pt] at (10.0,-7.5) {\includegraphics[width=2.5cm,height=2.5cm]{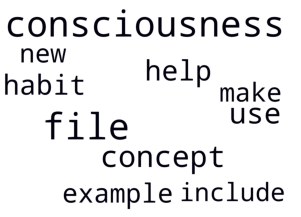}};
\draw[draw=black,line width=0.1pt] (10.0,-10.0) -- (12.5,-10.0);
\node[anchor=north west,inner sep=0pt] at (0.0,-10.0) {\includegraphics[width=2.5cm,height=2.5cm]{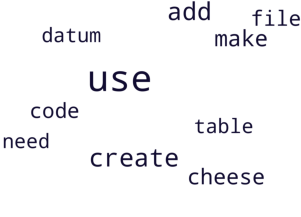}};
\draw[draw=black,line width=0.1pt] (2.5,-10.0) -- (2.5,-12.5);
\draw[draw=black,line width=0.1pt] (0.0,-12.5) -- (2.5,-12.5);
\node[anchor=north west,inner sep=0pt] at (2.5,-10.0) {\includegraphics[width=2.5cm,height=2.5cm]{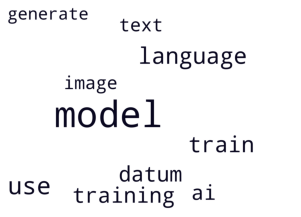}};
\draw[draw=black,line width=0.1pt] (5.0,-10.0) -- (5.0,-12.5);
\draw[draw=black,line width=0.1pt] (2.5,-12.5) -- (5.0,-12.5);
\node[anchor=north west,inner sep=0pt] at (5.0,-10.0) {\includegraphics[width=2.5cm,height=2.5cm]{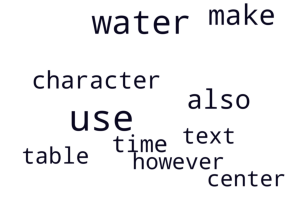}};
\draw[draw=black,line width=0.1pt] (7.5,-10.0) -- (7.5,-12.5);
\draw[draw=black,line width=0.1pt] (5.0,-12.5) -- (7.5,-12.5);
\node[anchor=north west,inner sep=0pt] at (7.5,-10.0) {\includegraphics[width=2.5cm,height=2.5cm]{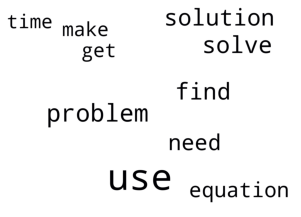}};
\draw[draw=black,line width=0.1pt] (10.0,-10.0) -- (10.0,-12.5);
\draw[draw=black,line width=0.1pt] (7.5,-12.5) -- (10.0,-12.5);
\node[anchor=north west,inner sep=0pt] at (10.0,-10.0) {\includegraphics[width=2.5cm,height=2.5cm]{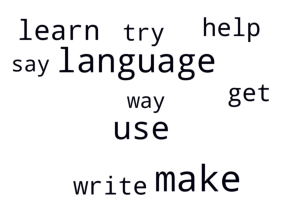}};
\draw[draw=black,line width=0.1pt] (10.0,-12.5) -- (12.5,-12.5);
\node[anchor=north west,inner sep=0pt] at (0.0,-12.5) {\includegraphics[width=2.5cm,height=2.5cm]{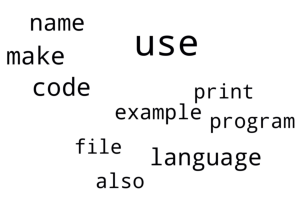}};
\draw[draw=black,line width=0.1pt] (2.5,-12.5) -- (2.5,-15.0);
\draw[draw=black,line width=0.1pt] (0.0,-15.0) -- (2.5,-15.0);
\node[anchor=north west,inner sep=0pt] at (2.5,-12.5) {\includegraphics[width=2.5cm,height=2.5cm]{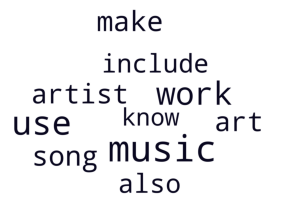}};
\draw[draw=black,line width=0.1pt] (5.0,-12.5) -- (5.0,-15.0);
\draw[draw=black,line width=0.1pt] (2.5,-15.0) -- (5.0,-15.0);
\node[anchor=north west,inner sep=0pt] at (5.0,-12.5) {\includegraphics[width=2.5cm,height=2.5cm]{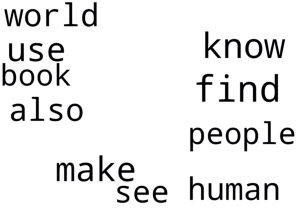}};
\draw[draw=black,line width=0.1pt] (7.5,-12.5) -- (7.5,-15.0);
\draw[draw=black,line width=0.1pt] (5.0,-15.0) -- (7.5,-15.0);
\node[anchor=north west,inner sep=0pt] at (7.5,-12.5) {\includegraphics[width=2.5cm,height=2.5cm]{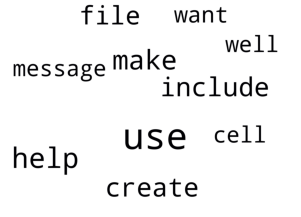}};
\draw[draw=black,line width=0.1pt] (10.0,-12.5) -- (10.0,-15.0);
\draw[draw=black,line width=0.1pt] (7.5,-15.0) -- (10.0,-15.0);
\node[anchor=north west,inner sep=0pt] at (10.0,-12.5) {\includegraphics[width=2.5cm,height=2.5cm]{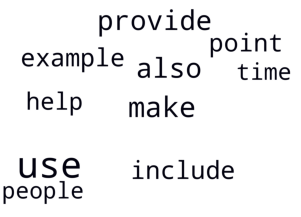}};
\draw[draw=black,line width=0.1pt] (10.0,-15.0) -- (12.5,-15.0);
\node[anchor=north west,inner sep=0pt] at (0.0,-15.0) {\includegraphics[width=2.5cm,height=2.5cm]{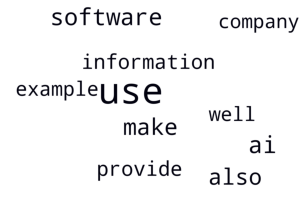}};
\draw[draw=black,line width=0.1pt] (2.5,-15.0) -- (2.5,-17.5);
\draw[draw=black,line width=0.1pt] (0.0,-17.5) -- (2.5,-17.5);
\node[anchor=north west,inner sep=0pt] at (2.5,-15.0) {\includegraphics[width=2.5cm,height=2.5cm]{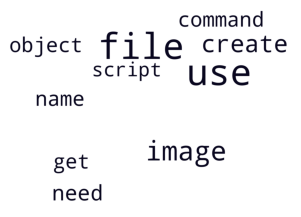}};
\draw[draw=black,line width=0.1pt] (5.0,-15.0) -- (5.0,-17.5);
\draw[draw=black,line width=0.1pt] (2.5,-17.5) -- (5.0,-17.5);
\node[anchor=north west,inner sep=0pt] at (5.0,-15.0) {\includegraphics[width=2.5cm,height=2.5cm]{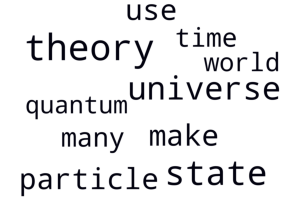}};
\draw[draw=black,line width=0.1pt] (7.5,-15.0) -- (7.5,-17.5);
\draw[draw=black,line width=0.1pt] (5.0,-17.5) -- (7.5,-17.5);
\end{tikzpicture}
    \caption{Word clouds for 33 topics extracted from the English subset of the open-assistant dataset.
    The number of topics was selected by selecting the peak coherence from 40 LDA models.
    One can observe high topic variety from biology (second top row), cooking (top row), and music (second to last row), all the way up to quantum physics (bottom row).
    }
    \label{fig:word_cloud}
\end{figure}

We present some statistics on the words used the most in Fig.~\ref{fig:word_cloud}.

\FloatBarrier
\section{Collection parameters}
\label{app:parameters}

\begin{table}[!htb]
    \centering
    \begin{tabular}{ll}\toprule
        Parameter & value\\\midrule
        max active trees &  100\\
        max initial prompt review & 100\\
        max tree depth & 5\\
        max children count & 2\\
        num prompter replies & 1\\
        goal tree size & 9\\
        %random goal tree size & false\\ not used, so not mentioned
        %min goal tree size & 5\\
        num reviews initial prompt & 3\\
        num reviews reply & 3\\
        auto mod enabled & true\\
        auto mod max skip reply & 25\\
        auto mod red flags & 4\\
        p full labeling review prompt & 1\\
        p full labeling review reply assistant & 1\\
        p full labeling review reply prompter & 0.1\\
        acceptance threshold initial prompt & 0.6\\
        acceptance threshold reply & 0.6\\
        num required rankings & 3\\
        p activate backlog tree & 0.1\\
        min active rankings per lang & 20\\
        lonely children count & 2 \\
        p lonely child extension & 0.75\\
        recent tasks span sec & 300\\
        max pending tasks per user & 8\\
        max prompt lottery waiting & 1000\\
   \bottomrule
    \end{tabular}
    \caption{Collection parameters}
    \label{tab:data_collection_parameters}
\end{table}
We provide more information on the parameters used to collect new data samples.
\begin{list}{}{\leftmargin=0em \itemindent=0em}
    \item \emph{max active trees}: Maximum number of concurrently active message trees in the database. No new initial prompt tasks are handed out to users if this number is reached
    \item \emph{max initial prompt review}: Maximum number of initial prompts under review before no more initial prompt tasks will be handed out.
    \item \emph{max tree depth}: Maximum depth of message tree.
    \item \emph{max children count}: Maximum number of reply messages per tree node.
    \item \emph{num prompter replies}: Number of prompter replies to collect per assistant reply.
    \item \emph{goal tree size}: Total number of messages to gather per tree.
    \item \emph{num reviews initial prompt}: Number of peer-review checks to collect in the `INITIAL\_PROMPT\_REVIEW` state
    \item \emph{num reviews reply}: Number of peer review checks to collect per reply (other than initial prompt).
    \item \emph{auto mod enabled}: Flag to enable/disable auto moderation.
    \item \emph{auto mod max skip reply}: Automatically set tree state to `halted\_by\_moderator` when more than the specified number of users skip replying to a message. (auto moderation)
    \item \emph{auto mod red flags}: Delete messages that receive more than this number of red flags if it is a reply or set the tree to `aborted\_low\_grade` when a prompt is flagged. (auto moderation)
    \item \emph{p full labeling review prompt}: Probability of full text-labeling (instead of mandatory only) for initial prompts.
    \item \emph{p full labeling review reply assistant}: Probability of full text-labeling (instead of mandatory only) for assistant replies.
    \item \emph{p full labeling review reply prompter}: Probability of full text-labeling (instead of mandatory only) for prompter replies.
    \item \emph{acceptance threshold initial prompt}: Threshold for accepting an initial prompt.
    \item \emph{acceptance threshold reply}: Threshold for accepting a reply.
    \item \emph{num required rankings}: Number of rankings in which the message participated.
    \item \emph{p activate backlog tree}: Probability to activate a message tree in BACKLOG\_RANKING state when another tree enters a terminal state.
    \item \emph{min active rankings per lang}: When the number of active ranking tasks is below this value when a tree enters a terminal state an available trees in BACKLOG\_RANKING will be activated (i.e. enters the RANKING state).
    \item \emph{lonely children count}: Number of children below which parents are preferred during sampling for reply tasks.
    \item \emph{recent tasks span sec}: Time in seconds of recent tasks to consider for exclusion during task selection.
    \item \emph{max pending tasks per user}: Maximum number of pending tasks (neither canceled nor completed) by a single user within the time span defined by `recent\_tasks\_span\_sec`.
    \item \emph{max prompt lottery waiting}: Maximum number of prompts in prompt\_lottery\_waiting state per language. If this value is exceeded no new initial prompt tasks for that language are generated.
\end{list}

\section{Training Configuration}
\label{sec:training_config}

Following~\cite{ouyang2022training} and as introduced in Section~\ref{sec:introduction}, we train supervised fine-tuned models (\textit{SFT}), reward models (RM), and a PPO fine-tuned models based on RM's predictions. We use as base models the popular decoder-only Pythia~\cite{biderman2023pythia} and LLaMA~\cite{touvron2023llama}.

\paragraph{Conversation format}

We sample threads in the CTs and provide them as input text to the model by using some additional special tokens. More specifically, a thread composed of prompts (P) $\text{P}_1, \text{P}_2, \dots $ and replies (R) $\text{R}_1, \text{R}_2, \dots$ is provided as input to the model with the following format:

\begin{align*}
    &\text{<prompter\_token>} \, \text{P}_1 \, \text{<endoftext\_token>} \, \text{<assistant\_token>} \, \text{R}_1 \, \text{<endoftext\_token>} \\   
    &\text{<prompter\_token>} \, \text{P}_2 \, \text{<endoftext\_token>} \, \text{<assistant\_token>} \, \text{R}_2 \, \text{<endoftext\_token>} \\   
    & \,\,\,\, \dots
\end{align*}
Each of the prompts and the replies consists potentially of multiple tokens after tokenizing.

\paragraph{\emph{SFT-mix} details.}
The \emph{sft-mix} training data configuration used for training \emph{OpenAssistant/falcon-40b-sft-mix-1226} is a mixture of OASST1 with other instruction tuning datasets, according to the following configuration:
\begin{lstlisting}
sft9-stage2:
  # oasst_export: 100.00% (29899)
  # vicuna: 50.00% (16963)
  # code_alpaca: 50.00% (9510)
  # oa_wiki_qa_bart_10000row: 100.00% (9434)
  # grade_school_math_instructions: 100.00% (8351)
  # dolly15k: 100.00% (14250)

  use_custom_sampler: true
  datasets:
    - oasst_export:
        lang: "bg,ca,cs,da,de,en,es,fr,hr,hu,it,nl,pl,pt,ro,ru,sl,sr,sv,uk" # sft-8.0
        input_file_path: 2023-06-02_oasst_all_labels.jsonl.gz
        val_split: 0.05
        top_k: 2
    - vicuna:
        fraction: 0.5
        val_split: 0.025
        max_val_set: 250
    - code_alpaca:
        fraction: 0.5
        val_split: 0.05
        max_val_set: 250
    - oa_wiki_qa_bart_10000row:
        val_split: 0.05
        max_val_set: 250
    - grade_school_math_instructions:
        val_split: 0.05
    - dolly15k:
        val_split: 0.05
        max_val_set: 300

\end{lstlisting}
More details can be found on the Hugging Face hub and in our open-source codebase.

\paragraph{Supervised fine-tuning.} During this phase, we fine-tune pretrained models for the regular language modelling tasks based on our conversational data. We mask tokens that correspond to prompts and only train to predict tokens that correspond to assistant replies. 

\paragraph{Reward model.} For the reward model training, we replace the language modelling head with a linear layer producing a single output $r_{\theta}$, corresponding to the predicted score for the last reply of the conversation. We use replies to the same prompt and their rankings as described in Appendix~\ref{app:ranking}.
Following~\cite{ouyang2022training}, assuming $K$ distinct replies, we produce ${K \choose 2}$ comparisons and train to minimize the loss
\begin{align*}
    \text{loss}(\theta) = - \frac{1}{{K \choose 2}} E_{(x, y_w, y_l)} [\log (\sigma(r_{\theta} (x, y_w) - r_{\theta}(x, y_l)))],
\end{align*}
where $\sigma$ is the sigmoid function and $y_w$ corresponds to a preferred completion for the pair of $y_w$ and $y_r$. We also optionally add another regularization parameter that prevents the predicted values from diverging too much.
Performance is measured by measuring the ability to predict the better reply among pairs of replies with different rank, on a held-out validation set.

\paragraph{PPO training.} We fine-tune the \textit{SFT} model by producing assistant replies to unanswered questions. We use the RM to score these replies and train with PPO, using the trlx framework~\footnote{\url{https://github.com/CarperAI/trlx}}. Following~\cite{ouyang2022training}, we also add a per-token KL penalty from the \textit{SFT} model at each token to avoid instability and over-optimization to the RM model.

All details and current training parameters are publicly available under \url{https://github.com/LAION-AI/Open-Assistant/tree/main/model/model_training}.

\section{Evaluation Tasks}
This section aims to provide a brief explanation of the evaluation tasks used in Table~\ref{tab:llm_evals}.

\emph{lm-evaluation-harness~\cite{eval-harness}} is a framework for evaluating language models on a variety of standardized tasks. We focus on the subset of BoolQ, PIQA, HellaSwag, WinoGrande, ARC-e, ARC-c and OBQA.

\emph{Vicuna Elo Rank~\cite{vicuna}} leverages LLMs to judge other LLMs and assign them a ranking according to the ELO system.

\emph{OpenAI Evals~\cite{openai2023gpt4}} is a set of general LLM evaluation benchmarks OpenAI released along with the announcement of GPT-4, and to which OpenAI has asked the community to contribute.

\emph{HumanEval~\cite{chen2021codex}} is a benchmark OpenAI released along with the announcement of Codex and sets out to measure "measure functional correctness for synthesizing programs from docstrings".

\FloatBarrier
\section{Political Compass Evaluations}
\label{app:political_compass}
% chatgpt: https://www.mdpi.com/2076-0760/12/3/148
% OA: https://docs.google.com/document/d/1tRQwPpGSGGzzRzC2oX2CXx1b-ZhmlvDXN5ruif55Shw/edit#

\begin{figure}[!htb]
    \centering
    \includegraphics[width=\textwidth]{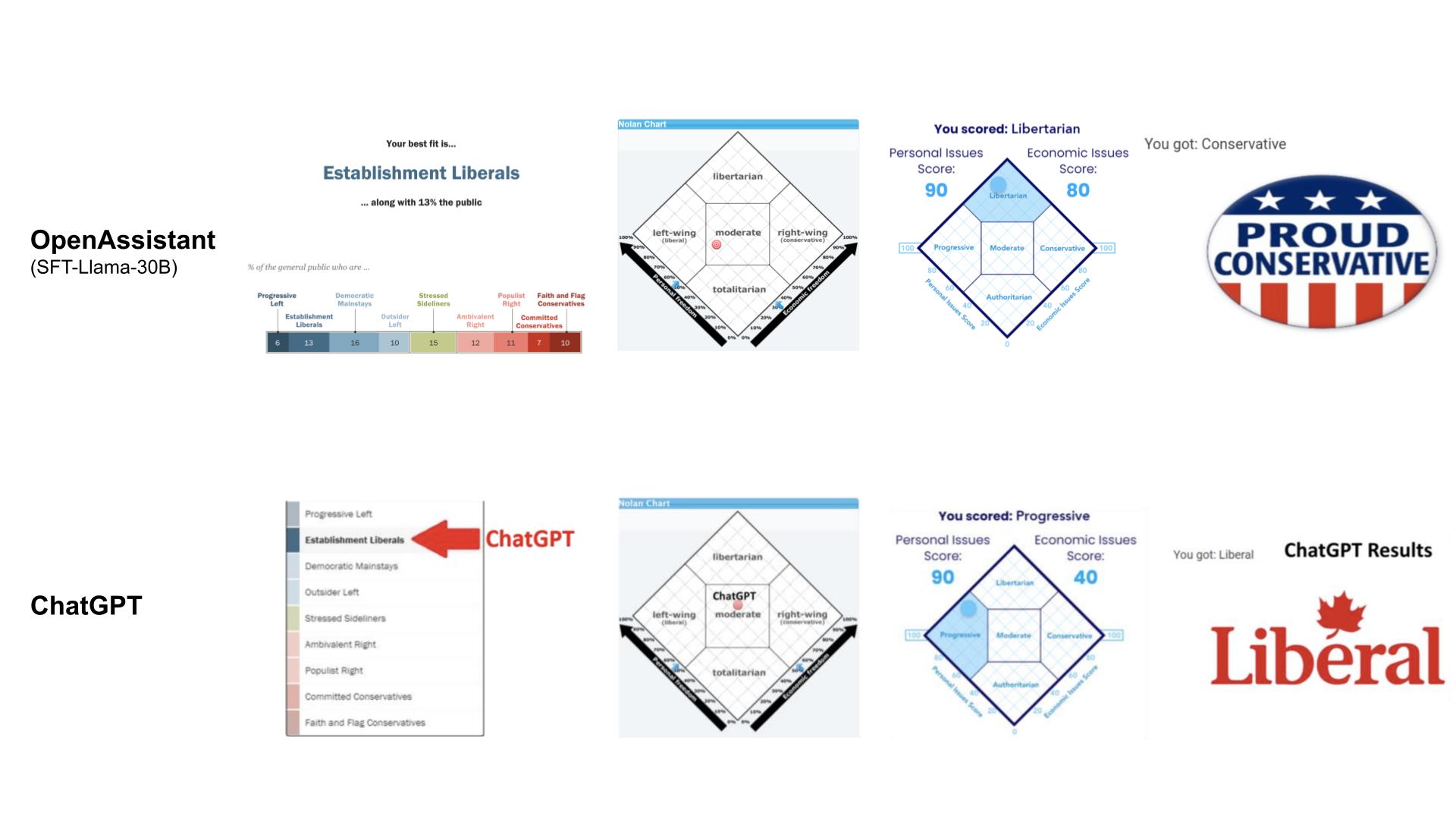}
    \caption{Comparison of evaluations on test for political leanings. For original ChatGPT results and references to tests used, see \cite{chatgptbias2023}}
    \label{fig:oa_bias}
\end{figure}

The political leanings of ChatGPT have been investigated in \cite{chatgptbias2023}.
We evaluated a model fine-tuned on OpenAssistant Conversations on a subset of the given tests.
Prompts were standardized and multiple samples were drawn, with majority vote deciding on the final answer for each question.
Figure~\ref{fig:oa_bias} depicts the result.
We stress that these are very preliminary results and should not be taken with large degrees of certainty, as the community has yet to find consensus on the exact methodology to perform such evaluations.
We will update this section with improved results in the future.
Our limited, preliminary results show the model trained on OpenAssistant Conversations to be more balanced and varied in its political leanings than ChatGPT.

\FloatBarrier
\section{Community Engagement}

Throughout the collection of OpenAssistant Conversations, a large global community has been built, including an active Discord group, and a GitHub repository with over 200 contributors.

\begin{figure}
    \includegraphics[width=\textwidth]{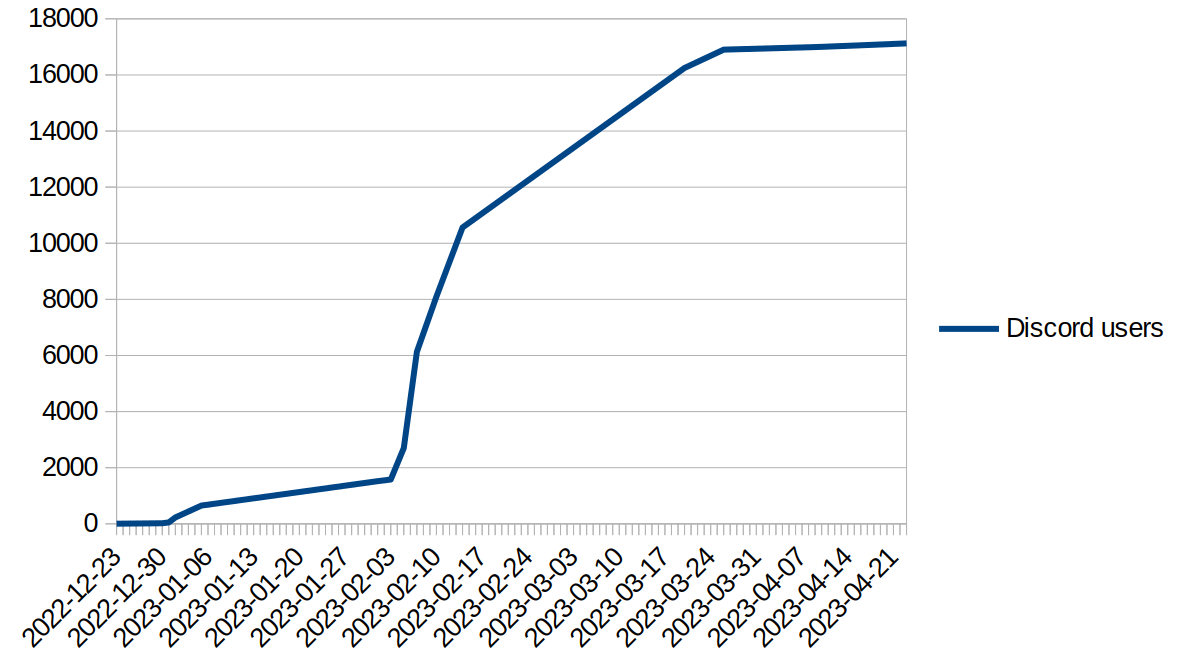}
    \caption{Discord users in the OpenAssistant group over time.}
    \label{fig:discord_users}
\end{figure}

Figure~\ref{fig:discord_users} shows the growth of the Discord community throughout the duration of data collection.

\begin{figure}
    \includegraphics[width=\textwidth]{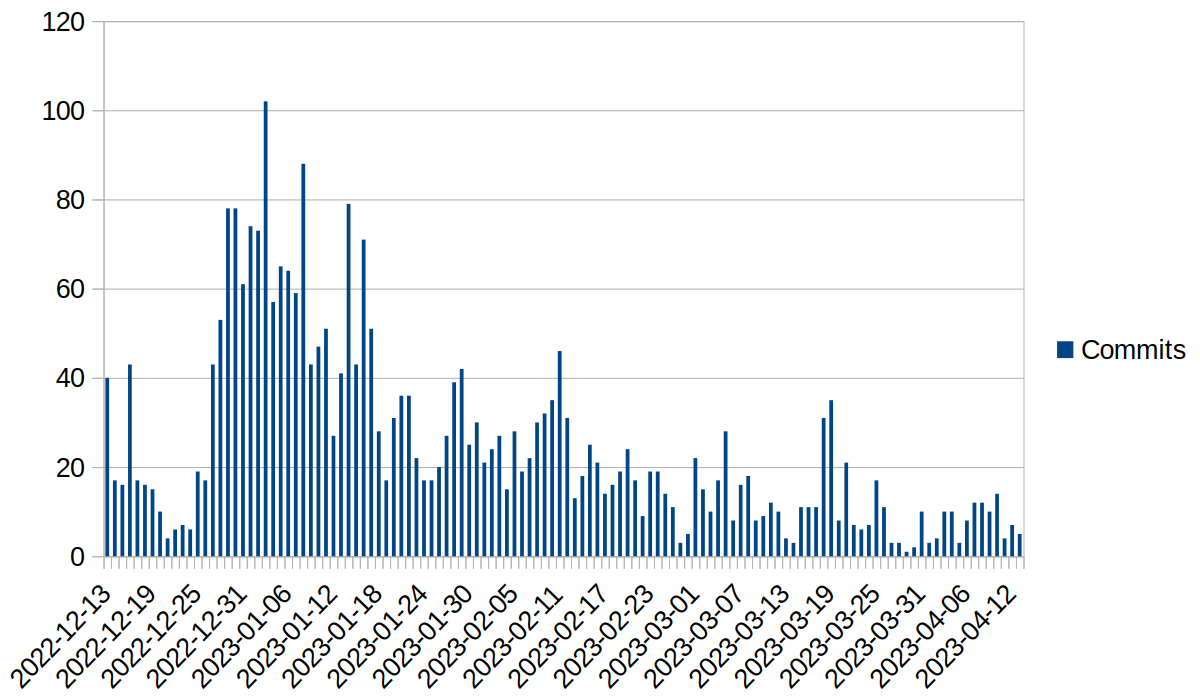}
    \caption{GitHub commits to the OpenAssistant repository over time.}
    \label{fig:github_commits}
\end{figure}

Figure~\ref{fig:github_commits} shows new commits to the GitHub repository over time.

\begin{figure}
    \includegraphics[width=\textwidth]{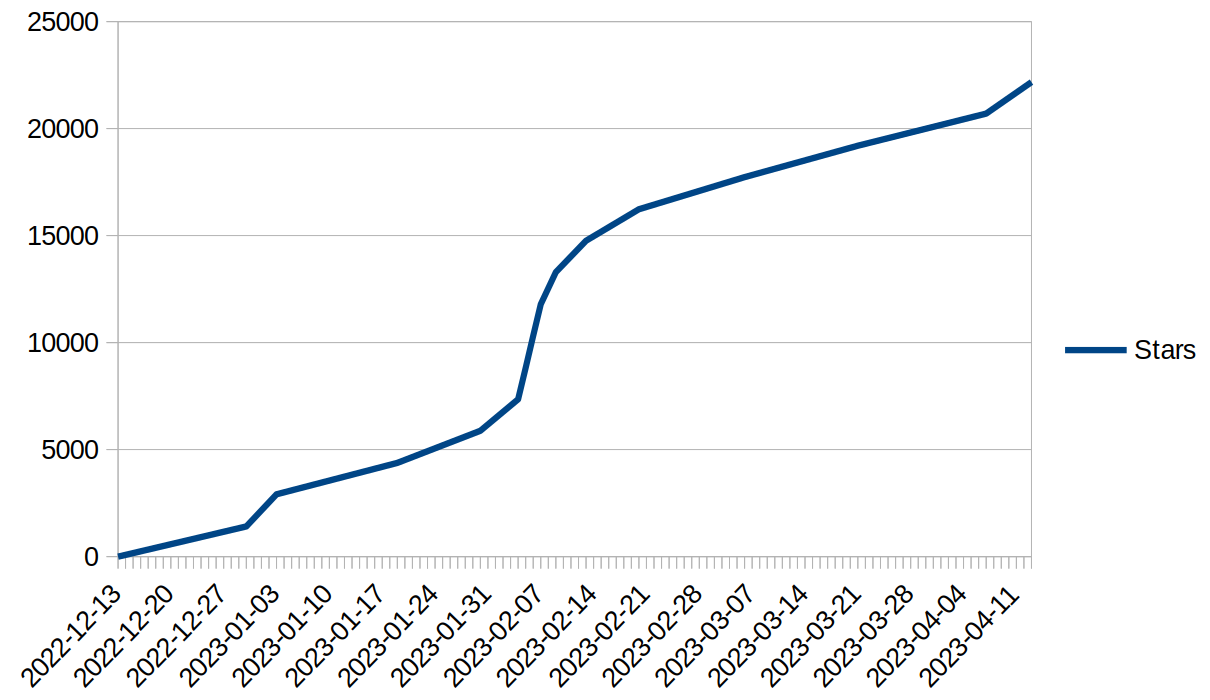}
    \caption{GitHub stars on the OpenAssistant repository over time.}
    \label{fig:github_stars}
\end{figure}

Figure~\ref{fig:github_stars} shows the growth in stars on the GitHub repository over time.

\begin{figure}
    \includegraphics[width=\textwidth]{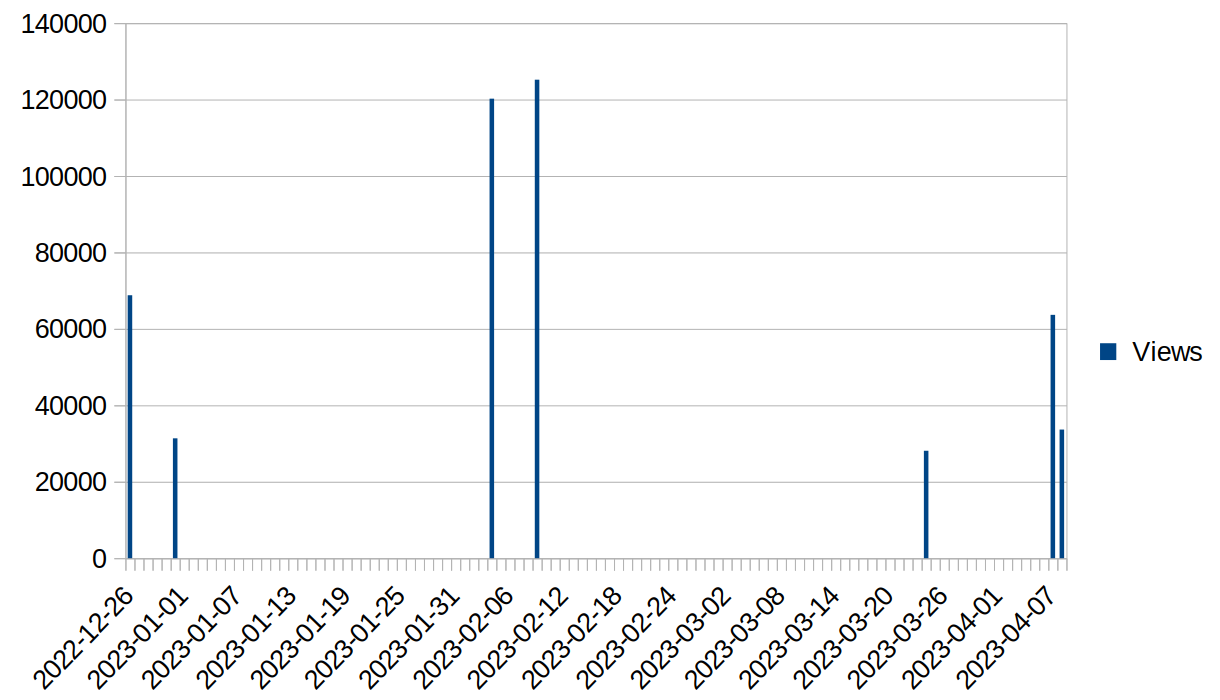}
    \caption{YouTube's videos' views on the OpenAssistant theme over time.}
    \label{fig:video_views}
\end{figure}

Figure~\ref{fig:video_views} shows popularity of OpenAssistant by YouTube's videos' views on the theme over time.

These Figures serve as a strong reminder of what can be achieved by the collective effort of many volunteers, even in a field where research has thus far been largely monopolized by a small number of industrial labs.

In addition, by comparing the massive influx of new contributors and subscribers to the emergence of Open Assistant themed videos, it shows how certain media events have influenced the development.

\newpage
\section{Dataset Documentation (Data card)}

Dataset documentation can be found at \data.

\section{Author Statement}

We confirm that we bear all responsibility in case of any violation of rights during the collection of the data or other work, and will take appropriate action when needed, e.g. to remove data with such issues.

\end{document}